\documentclass[lettersize,journal]{IEEEtran}

\usepackage{times}
\usepackage{epsfig}
\usepackage{graphicx}
\usepackage{amsmath}
\usepackage{amssymb}
\usepackage{stfloats}
\usepackage{multirow}
\usepackage{bbm}
\usepackage{rotating}
\usepackage{array}
\usepackage{color}
\usepackage{threeparttable}

\usepackage{epstopdf}
\usepackage{pifont}
\usepackage{subfigure}
\usepackage{bm}
\usepackage{url}
\usepackage[table]{xcolor}
\usepackage{color}
\usepackage{booktabs}
\usepackage{dsfont}
\usepackage[linesnumbered,ruled,vlined]{algorithm2e}

\begin{document}

\title{Unsupervised Attention Regularization Based Domain Adaptation for Oracle Character Recognition}

\author{Mei Wang, Weihong Deng, Jiani Hu, Sen Su
\thanks{Mei Wang, Weihong Deng and Jiani Hu are with the Pattern Recognition and Intelligent System Laboratory, School of Artificial Intelligence, Beijing University of Posts and Telecommunications, Beijing, 100876, China. E-mail: \{wangmei1,whdeng,jnhu\}@bupt.edu.cn. (Corresponding Author: Weihong Deng)}
\thanks{Sen Su is with State Key Laboratory of Networking and Switching Technology, Beijing University of Posts and Telecommunications, Beijing, 100876, China. E-mail: susen@bupt.edu.cn.}}

\markboth{Journal of \LaTeX\ Class Files,~Vol.~14, No.~8, August~2021}%
{Shell \MakeLowercase{\textit{et al.}}: A Sample Article Using IEEEtran.cls for IEEE Journals}

\IEEEpubid{0000--0000/00\$00.00~\copyright~2021 IEEE}

\maketitle

\begin{abstract}
The study of oracle characters plays an important role in Chinese archaeology and philology. However, the difficulty of collecting and annotating real-world scanned oracle characters hinders the development of oracle character recognition. In this paper, we develop a novel unsupervised domain adaptation (UDA) method, i.e., unsupervised attention regularization network (UARN), to transfer recognition knowledge from labeled handprinted oracle characters to unlabeled scanned data. First, we experimentally prove that existing UDA methods are not always consistent with human priors and cannot achieve optimal performance on the target domain. For these oracle characters with flip-insensitivity and high inter-class similarity, model interpretations are not flip-consistent and class-separable. To tackle this challenge, we take into consideration visual perceptual plausibility when adapting. Specifically, our method enforces attention consistency between the original and flipped images to achieve the model robustness to flipping. Simultaneously, we constrain attention separability between the pseudo class and the most confusing class to improve the model discriminability. Extensive experiments demonstrate that UARN shows better interpretability and achieves state-of-the-art performance on Oracle-241 dataset, substantially outperforming the previously structure-texture separation network by 8.5\%.
\end{abstract}


\section{Introduction}

\IEEEPARstart{O}{racle} characters \cite{flad2008divination,keightley1997graphs} are the oldest hieroglyphs in China, which are engraved on tortoise shells and animal bones. They have far-reaching research value as treasures that recorded the ancient culture and history of the Shang Dynasty (around 1600-1046 B.C.). To help archaeologists and paleographists with the recognition of oracle characters, deep convolutional neural networks (CNN) \cite{he2016deep} are recently introduced \cite{huang2019obc306,zhang2019oracle}. While these deep models excel at capturing complex and hierarchical patterns from a sufficiently large dataset, it is challenging in practice to collect enough labeled oracle data. Real-world scanned oracle characters are extremely scarce, and the annotation process is expensive and time-consuming even for experts. One alternative that could mitigate this constraint is to leverage handprinted oracle characters, which are easy to acquire and annotate. However, the model trained with handprinted data often experiences a performance drop when applied to real-world scanned data due to domain discrepancy. Unsupervised domain adaptation (UDA) \cite{wang2018deep,liu2022deep} has emerged as a vital solution for this issue by transferring knowledge from a label-rich source domain (handprinted oracle data) to an unlabeled target domain (scanned characters).

\begin{figure}
\centering
\includegraphics[width=8.5cm]{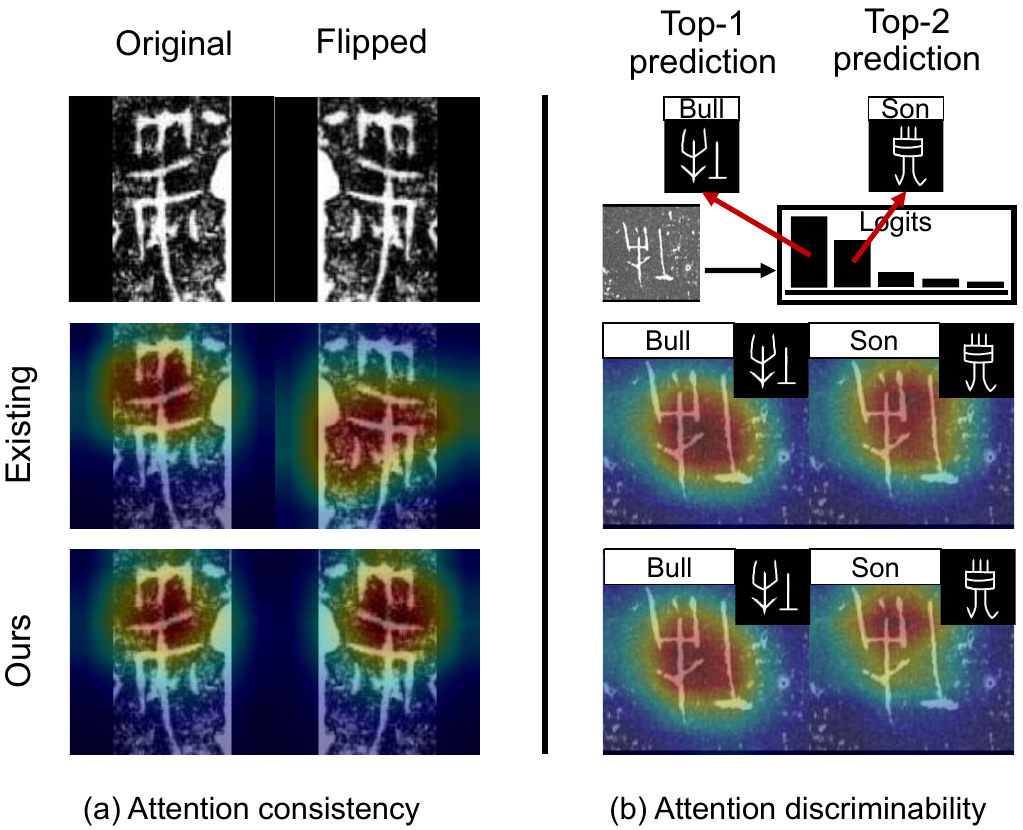}
\caption{An illustration of attention maps for scanned oracle characters. (a) Flipping an image does not flip the attention map in the existing UDA method, while our UARN significantly improves attention consistency. (b) Existing UDA method pays attention to similar regions when regarding the relevant pixels for classes “bull”  and “son”, while our UARN makes the attention map separable and tells the confusing class apart on the target domain. }
\label{overview} 
\end{figure}

Conventional UDA methods in other tasks mainly focus on reducing the distribution discrepancy between domains via moment matching \cite{Long2015Learning,Long2016Unsupervised} or adversarial learning \cite{Ganin2015Unsupervised}. \IEEEpubidadjcol
However, there are two limitations when these methods are directly applied to oracle character recognition. First, different from other characters, oracle characters are pictographic and even flip-insensitive. Therefore, the learned models should be robust to flipping. However, we find that existing UDA methods often \emph{fail to preserve interpretation consistency under this spatial transformation on the target domain}. For example, flipping a target image horizontally does not flip the attention heatmap \cite{zhou2016learning, selvaraju2017grad}, even if the random-flip augmentation is performed on training data, as shown in Fig. \ref{overview}(a). Second, in addition to different writing styles, oracle characters belonging to the same category largely vary in stroke and even topology, which increases the intra-class variations. The inter-class similarity of characters is also extremely disturbing to the performance. Thus, discriminability is crucial to oracle character recognition. However, alignment-based UDA methods \cite{Long2015Learning,Long2016Unsupervised,Ganin2015Unsupervised} \emph{suffer from confused interpretations across different target classes.} That is to say, the attention heatmaps corresponding to an individual class of interest are not discriminative across classes. As shown in Fig. \ref{overview}(b), there are large overlapping regions between the attention map of the top-1 prediction and that of the top-2 prediction. The inconsistent and inseparable attention heatmaps would result in worse visual perceptual plausibility and sub-optimal performance on the target domain when adapting.

In this paper, we take interpretability into account when adapting and propose a novel UDA method for oracle character recognition, called unsupervised attention regularization network (UARN), which incorporates attention consistency and discriminability in the adapting process. To provide visual explanations for the model’s predictions, we use class activation mapping (CAM) \cite{zhou2016learning} to generate the attention heatmap for each class on the corresponding target image. For attention consistency, we assume that the learned attention heatmaps should follow the same transformation as the input images to achieve the model robustness on the target domain. Therefore, our UARN reduces the distance between the attention map of the flipped image and the flipped attention map of the original image. We encourage the consistency on the attention maps of all classes, not just the ground-truth class, which enforces a stricter constraint and bypasses the demand of target ground-truth labels. For attention discriminability, our intuition is that the spatial regions that most contributed to the output in a given feature map should be different across target classes such that the model discriminability is improved and visual confusion is reduced. To this end, our UARN makes the attention heatmap of the ground-truth class and that of the most confusing class separable on the target domain to tell the confusing class apart. To address the problem of lacking ground-truth labels on the target domain, we introduce pseudo-labeling \cite{saito2017asymmetric,liu2021cycle} and take the pseudo class with high confidence as a substitute for the ground-truth class. Experiments show that our UARN improves the consistency and separability of attention maps as well as the classification accuracy on scanned oracle characters.

Our contributions can be summarized into three aspects.

1) Oracle character recognition is still an understudied field of research. We propose a novel UDA method, i.e., unsupervised attention regularization network, to improve the model performance on real-world scanned data, which contributes not only to technology but also to the understanding of ancient civilization.

2) Our proposed UARN takes interpretability into consideration and encourages better visual perceptual plausibility when adapting. Attention consistency improves the model robustness to flipping on the target domain, and simultaneously, attention discriminability reduces visual confusion across different target classes.

3) Extensive experimental results on Oracle-241 dataset demonstrate that our method shows better interpretability and successfully transfers recognition knowledge from handprinted oracle characters to scanned data. It substantially outperforms the recently proposed structure-texture separation network \cite{9757826} by 8.5\%.

\section{Related work}

\subsection{Oracle character recognition}

Oracle character recognition aims to classify characters from drawn or real rubbing oracle bone images. Earliest works primarily leveraged graph theory and topology to extract hand-crafted features and perform recognition. Gu et al. \cite{shaotong2016identification} recognized characters based on topological registration. Lv et al. \cite{lv2010graphic} utilized a Fourier descriptor based on curvature histograms to represent oracle data. Li et al. \cite{li2011recognition} regarded each oracle character as an undirected graph, and classified it by graph isomorphism. 
Guo et al. \cite{guo2015building} constructed an Oracle-20K dataset for handprinted oracle characters and proposed hierarchical representations combining Gabor-related and sparse encoder-related features.

To address the limitation of hand-crafted features, CNNs are recently introduced and facilitate the development of oracle character recognition. Huang et al. \cite{huang2019obc306} constructed an OBC-306 dataset for scanned oracle data, and trained AlexNet \cite{krizhevsky2012imagenet}, VGGNet \cite{simonyan2014very} and ResNet \cite{he2016deep} to perform recognition. Lin et al. \cite{lin2022radical} integrated the convolutional block attention module (CBAM) \cite{woo2018cbam} into deep network to detect the radicals of oracle characters. Liu et al. \cite{liu2022one} proposed a siamese similarity network for one-shot oracle character recognition, which utilized the multi-scale fusion backbone and soft similarity contrast loss to improve the model’s ability. Li et al. \cite{li2021mix} introduced mixup augmentation to address the problem of imbalanced data distribution for oracle characters.

Although UDA can be one of the powerful approaches to address the problem of insufficient data for oracle character recognition, it is still an understudied field of research. To our knowledge, there is only one work \cite{9757826} focusing on it, which disentangled features into structure and texture components and further realized image-translation across domains. In this paper, we propose a simple and effective UDA method with the help of attention regularization.

\subsection{Unsupervised domain adaptation}

UDA \cite{wang2018deep} has been studied extensively in recent years, largely for alleviating data annotation constraint. A major line of
work aligns the source and target domains by minimizing a divergence that measures the discrepancy between domains, such as maximum mean discrepancy (MMD) \cite{long2018transferable, 9832602}, correlation alignment (CORAL) \cite{sun2016deep} and kullback-leiber divergence (KL) \cite{zhuang2015supervised}. For example, Zhang et al. \cite{8370105} minimizd a MMD-based class-wise fisher discriminant across domains to match the distribution for each class. CAN \cite{kang2019contrastive} simultaneously optimized the intra-class and inter-class domain discrepancy by a new metric established on MMD. HoMM \cite{chen2020homm} matched the third- and fourth-order statistics to perform fine-grained domain alignment.

Another promising direction is based on adversarial training \cite{Tzeng2017Adversarial,gallego2020incremental,9495801}, which learns invariant features by deceiving a domain discriminator. For example, Cui et al. \cite{cui2020gradually} equipped adversarial learning with gradually vanishing bridge (GVB) mechanism to reduce the negative influence of domain-specific characteristics. Xu et al. \cite{8946732} proposed an importance sampling method for adversarial domain adaptation to adaptively adjust the model gradient for each sample. Zuo et al. \cite{9435354} jointed adversarial domain adaptation with margin-based generative module to enhance the model discrimination.

Self-training (also called pseudo-labeling) \cite{saito2017asymmetric,wang2022adaptive} has also been applied in UDA to compensate for the lack of categorical information on the target domain. Deng et al. \cite{8964455} applied a classifier to generate pseudo-labels for target data, and then performed class-level alignment via triplet loss. Sun et al. \cite{sun2022prior} proposed to refine pseudo labels using prior knowledge. Wang et al. \cite{wang2020unsupervised} proposed a novel selective pseudo-labeling strategy based on structured prediction and learned a domain invariant subspace by supervised locality preserving projection. Gu et al. \cite{gu2020spherical}  proposed a novel robust pseudo-label loss in spherical feature space for utilizing target pseudo-labels more robustly.

Different from the aforementioned work which focuses on UDA of object classification, we design a novel UDA method for oracle character recognition, incorporating attention regularization for enhanced target performance.

\begin{figure*}
\centering
\includegraphics[width=17.5cm]{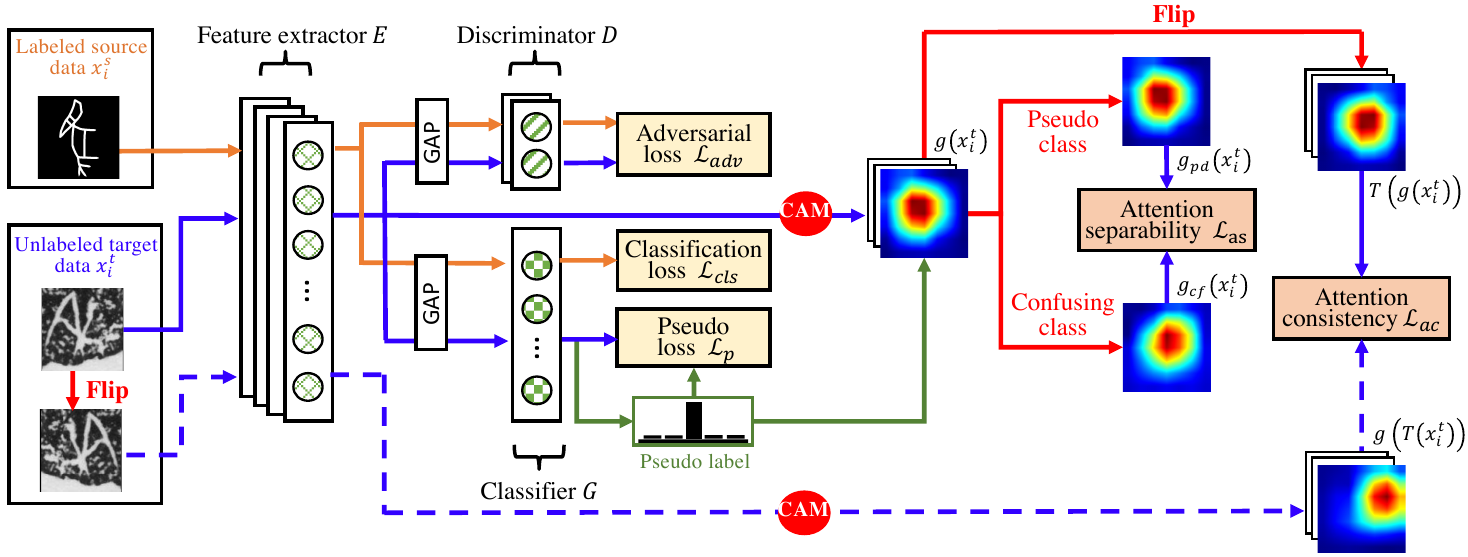}
\caption{Illustration of UARN. We utilize adversarial learning and pseudo labeling to learn domain-invariant features, and simultaneously enforce the consistency and discriminability of attention heatmaps to achieve better classification accuracy and visual perceptual plausibility on the target domain. For attention consistency, we reduce the distance between the attention map of the flipped image and the flipped attention map of the original image to improve the model robustness to flipping. For attention discriminability, we reduce the overlaps of attention maps between the pseudo class and the most confusing class to eliminate visual confusion.}
\label{architecture} 
\end{figure*}

\section{Methodology}

Following the settings of UDA, we define a \textbf{labeled} source domain $\mathcal{D}^s=\{{x^{s}_{i},y^{s}_{i}}\}^{N_s}_{i=1}$ of $N_s$ handprinted oracle characters, and an \textbf{unlabeled} target domain $\mathcal{D}^t=\{{x^{t}_{i}}\}^{N_t}_{i=1}$ of $N_t$ scanned oracle characters. Source and target domains share the same label space, and $K$ denotes the number of classes. Each oracle character locates in an image. Handprinted oracle characters are written by experts, while scanned oracle data are generated by reproducing the oracle-bone surface. Thus, the discrepancy between these two domains raises the key technical challenge of domain adaptation. Our goal is to learn a function $f$ using $\{{x^s_i,y^{s}_{i}}\}^{N_s}_{i=1}$ and $\{{x^{t}_{i}}\}^{N_t}_{i=1}$ which can classify the unlabeled target dataset without accessing the corresponding labels, in spite of the large domain discrepancy.

\subsection{Overview}

The overall framework of our proposed UARN is shown in Fig. \ref{architecture}. The model consists of a feature extractor $E$, a classifier $G$ and a discriminator $D$. Both source and target data ($x^s_i$ and $x^t_i$) first pass through the extractor $E$ to obtain the feature maps $F^{i,s},  F^{i,t} \in \mathbb{R}^{C\times H \times W}$ where $C$, $H$ and $W$ respectively represent the number of channels, height, width of the feature map. Global average pooling (GAP) is then applied on $F^{i,s}$ and $F^{i,t}$ to obtain feature vectors $z_i^s, z_i^t \in \mathbb{R}^{C\times 1}$, and finally the classifier $G$ is used to make the prediction. We train $E$ and $G$ by classification loss $\mathcal{L}_{cls}$ supervised with source labels. To achieve adaptation,  pseudo-labeling and adversarial learning are also performed.

Meanwhile, we flip target image $x^t_i$ to get its flipped counterpart $T(x^t_i)$, and fed $T(x^t_i)$ to $E$ to obtain its feature map. Then, the attention heatmaps $g(x_i^t)$ and $g\left(T\left(x_i^t\right)\right)$ are generated for $x^t_i$ and $T(x^t_i)$ via CAM. To enforce the consistency and separability of attention map, we minimize the distance between $g\left(T\left(x_i^t\right)\right)$ and the flipped version of $g(x_i^t)$, and reduce the overlaps of attention maps between the pseudo-class $g_{pd}(x_i^t)$ and the most confusing class $g_{cf}(x_i^t)$.

\subsection{Pseudo-labeling}

To learn a basic recognition model, we optimize the network on handprinted data in a supervised way:
\begin{equation}
\mathcal{L}_{cls}= \mathbb{E}_{\left ( x_i^s,y_i^s \right )\sim \mathcal{D}^s} L_{CE}\left ( f\left (x_i^s\right ),y_i^s \right ) , \label{source}
\end{equation}
where $f=G\circ \text{GAP} \circ E$ is the recognition model, and $L_{CE}$ is the cross-entropy loss. Considering that the label of target data $x^t_i$ is unavailable, we pick up the class with the maximum predicted probability as its pseudo label $\hat{y}_i^t$,
\begin{equation}
\hat{y}_i^t=\begin{cases}
 & \arg\max f\left (x_i^t\right ), \ \ \ \max f\left (x_i^t\right )> \tau, \\
 &-1, \ \ \ \ \ \ \ \ \ \ \ \ \ \ \ \ \text{ otherwise. }  \label{pseudo}
\end{cases}
\end{equation}
To mask out noisy unlabeled samples, we only assign pseudo labels to high-confident data cut off by a pre-defined threshold $\tau$. After generating pseudo labels, the model can be simultaneously optimized on scanned oracle characters:
\begin{equation}
\mathcal{L}_{p}= \mathbb{E}_{ x_i^t \sim \mathcal{D}^t}\mathds{1}\left ( \max f\left (x_i^t\right )> \tau \right )  L_{CE}\left ( f\left (x_i^t  \right ),\hat{y}_i^t \right ). \label{target}
\end{equation}

\subsection{Adversarial learning}

Due to the distribution discrepancy between the two domains, the model suffers from performance degradation when applied directly to the scanned domain. Therefore, we apply adversarial learning following \cite{Ganin2015Unsupervised} to make the model invariant to domain-specific variations, thus aligning the distribution and improving its generalization across different domains. Specifically, it involves two main components: the extractor $E$ and the domain discriminator $D$. $D$ aims to distinguish features of samples from the two domains, while $E$ learns to confuse $D$,
\begin{equation}
\begin{split}
\underset{E}{min}\ \underset{D}{max} \ \mathcal{L}_{adv}&= \mathbb{E}_{x_i^s \sim \mathcal{D}^s} \log\left [ D\left ( \text{GAP}\left (E\left ( x_i^s \right )  \right )  \right ) \right ]\\
&+ \mathbb{E}_{x_i^t \sim \mathcal{D}^t} \log\left [ 1-D\left ( \text{GAP}\left (E\left ( x_i^t \right )  \right ) \right )  \right ]. \label{adversarial}
\end{split}
\end{equation}
This min-max game is expected to reach an equilibrium where features are domain-invariant.

\subsection{Attention consistency}

\begin{figure}
\centering
\includegraphics[width=8.5cm]{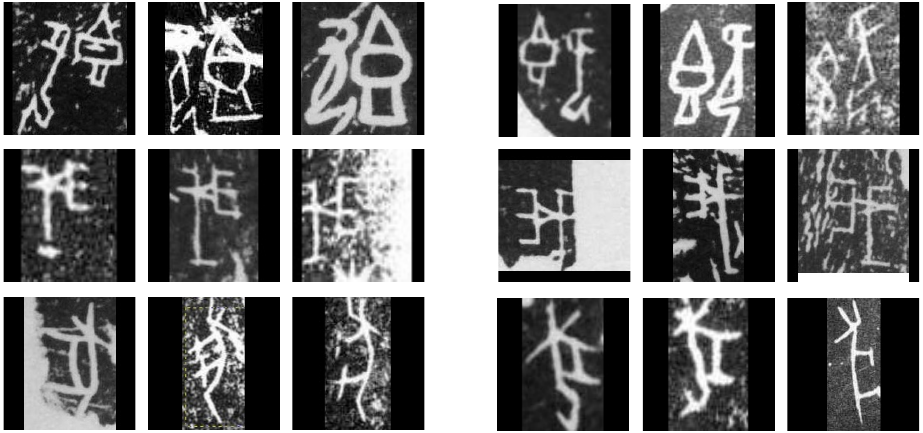}
\caption{The characters in the right three columns exhibit a left-right mirrored orientation in comparison to those in the left three columns.}
\label{mirror} 
\end{figure}


Different from other characters, oracle characters sometimes exhibit a left-right mirrored orientation compared to each other within the same category, as shown in Fig. \ref{mirror}. This phenomenon can be attributed to the pictographic nature of these characters. Therefore, all the characters can be flipped and it makes no difference for recognition from the perspective of human visual perception, i.e., the class labels. We hope the recognition model to exhibit the same level of robustness to the characters with left-right mirrored orientation as humans. However, we experimentally demonstrate that the existing adapted model is not robust enough to flipping, as shown in Fig. \ref{overview}(a). Specifically, when oracle characters are flipped, the model's attention heatmap undergoes a shift, indicating that the model relies on different regions for predictions. This observation challenges the conventional assumption that the model should employ consistent criteria for decision-making when recognizing identical characters, irrespective of their orientation. The inconsistency in decision-making criteria poses a risk of degrading performance on the target domain. To address this issue, we constrain that the attention heatmaps should be consistent before and after flipping the characters, i.e., reduce the distance between the attention map of the flipped image and the flipped attention map of the original image, and incorporate this regularization into the model training to enhance the model robustness and thus improve target performance.

\textbf{Class activation mapping} \cite{zhou2016learning}. CAM is utilized to visualize the input image regions used when CNN making decisions. We first utilize CAM to generate attention maps for each class on target images, which can be computed as:
\begin{equation}
\mathcal{A}_k=g_k( x ) = \sum_{j=1}^{C}  \omega_{kj}F_j, \label{attention}
\end{equation}
where $\mathcal{A}_k=g_k( x )\in \mathbb{R}^{H\times W}$ indicates the attention heatmap of image $x$ for class $k$. $C$ is the channel number of feature map. $F_j$ represents the $j$-th channel of feature map from the last convolutional layer. We denote the weight of the classifier as $W\in \mathbb{R}^{K\times C}$, and $\omega_{kj}$ represents the $(k,j)$ element of $W$ corresponding to the $k$-th class for the $j$-th channel of feature maps.

\textbf{Consistency regularization.} We feed a target image $x_i^t$ and its flipped counterpart $T(x^t_i)$ into extractor $E$, and compute their attention heatmaps $\mathcal{A}^{i,t}=g(x_i^t)$ and $\overline{\mathcal{A}^{i,t}} =g\left(T\left(x_i^t\right)\right)$, respectively. We omit the superscript $t$ of $\mathcal{A}$ in the following paragraphs for brief.

Based on the definition of attention consistency, $\mathcal{A}^{i}$ and $\overline{\mathcal{A}^{i}}$ need to be equivariant under the flip transformation, i.e., $T\left(g\left(x_i^t\right)\right)=g\left(T\left(x_i^t\right)\right)$. Therefore, we use the consistency loss to minimize the distance between the attention map of the flipped image and the flipped attention map of the original image:
\begin{equation}
\mathcal{L}_{ac}= \frac{1}{N_tKHW}  \sum_{i=1}^{N_t}  \sum_{k=1}^{K} \mathds{1}\left ( \max p_i> \tau \right ) \left \| T\left ( \mathcal{A}^{i}_k \right )-\overline{ \mathcal{A}^{i}_k} \right \| _2, \label{consistency}
\end{equation}
where $p_i= f\left (x_i^t\right )$ is the predicted probability of $x_i^t$ and $T\left ( \cdot  \right )$ denotes the flip transformation. $N_t$ and $K$ represent the number of target images and categories. $H$ and $W$ denote the height and width of feature maps. Since low-confident samples would incur inaccurate attention maps when lacking target labels in UDA, we also mask them out using $\tau$.

\subsection{Attention discriminability}


Large intra-class variation and high inter-class similarity make it difficult for existing UDA methods to recognize scanned oracle characters. They often struggle to distinguish between various classes. For example, given a scanned oracle character as shown in Fig. \ref{overview}(b), the model classifies it as the class “bull” with the highest probability (top-1 prediction), and as the class “son” with the second-highest probability (top-2 prediction). However, when attention maps are generated for the “bull” and “son” categories, significant overlap is observed between them. This implies that the network considers the regions most relevant to the predictions of these two categories to be similar. It may result in the model neglecting the distinctive structures inherent to characters of different categories, thereby hindering its ability to effectively learn discriminative features and tell various classes apart. To mitigate visual confusion, the model should prioritize attending to distinct regions relevant to the unique structures of different classes when making decisions. Therefore, we take class-separable attention as a principled part of the model training and design an attention discriminative loss, which reduces the overlaps of attention maps between different classes, i.e., the ground-truth class and the most confusing class.

However, the ground-truth labels are unavailable for target data in UDA of oracle character recognition. To address this issue, we generate pseudo labels by Eq. (\ref{pseudo}) and take the pseudo class as a substitute for the ground-truth class. For each scanned data $x_i^t$, the attention discriminative loss can be formulated as,
\begin{equation}
\mathcal{L}_{as}^{i}= 2\frac{\sum_{(h,w)} \left (  \min\left ( \mathcal{A}^{i}_{pd}(h,w), \mathcal{A}^{i}_{cf}(h,w)\right ) \cdot \mathcal{M}^{i}(h,w) \right ) }{\sum_{(h,w)}\left (  \mathcal{A}^{i}_{pd}(h,w)+ \mathcal{A}^{i}_{cf}(h,w)\right ) }  ,
\end{equation}
where $\mathcal{A}^{i}_{pd}(h,w)$ is the attention heatmap of target data $x_i^t$ at spatial position $(h,w)$ for the pseudo class. Similarly, $\mathcal{A}^{i}_{cf}(h,w)$ is the attention heatmap of target data $x_i^t$ at spatial position $(h,w)$ for the most confusing class. The most confusing class can be obtained by picking up the class with the second largest predicted probability. $\mathcal{M}$ denotes the mask to ignore the noise from background pixels and focus more on the pixels from the foreground region,
\begin{equation}
\mathcal{M}^{i}(h,w)=\frac{1}{1+exp\left ( -\alpha\left (  \mathcal{A}^{i}_{pd}(h,w)-\beta \right )   \right ) } , \label{sepera8}
\end{equation}
where $\alpha$ and $\beta$ are empirically set to be 100 and $0.55\times \left (\max  \mathcal{A}^{i}_{pd}  \right )  $, respectively.

However, the learned model might be incapable of precisely assigning pseudo labels for scanned oracle characters when the domain discrepancy is large. The hard-to-transfer examples with inaccurate pseudo classes may deteriorate the optimization procedure of attention separability. To reduce the negative influence of these samples, we prioritize the class-separable attention on easy-to-transfer examples by reweighting each training example via an entropy-aware weight $\varphi \left ( H(p_i) \right )$,
\begin{equation}
\mathcal{L}_{as}= \frac{1}{N_t} \sum_{i=1}^{N_t}\mathds{1}\left (\max p_i> \tau \right ) \varphi \left ( H(p_i) \right ) \mathcal{L}_{as}^{i} \label{separable}
\end{equation}
where $H(\cdot )$ denotes the entropy, and $\varphi \left ( H(p_i) \right )=1+exp\left ( -H\left ( p_i \right )  \right ) $ measures the certainty of model prediction for $x_i^t$. According to the definition, our attention discriminative loss would emphasize more on target data with higher prediction confidences and assign larger weights to them.

\subsection{Overall objective}

Combining the classification loss, pseudo loss, adversarial loss, attention consistency and attention discriminability, our overall objective is formulated as:
\begin{equation}
\begin{split}
&\underset{E,G}{min}\ \mathcal{L}_{cls}+\mathcal{L}_{p}+ \mathcal{L}_{adv}+\mu \mathcal{L}_{ac}+\lambda \mathcal{L}_{as}, \\
&\underset{D}{max}\ \mathcal{L}_{adv},
\end{split}
\end{equation}
where $\mu$ and $\lambda$ are the trade-off parameters to balance losses. We maximize $\mathcal{L}_{adv}$ to optimize the discriminator, and simultaneously minimize other losses to optimize the feature extractor and classifier. $\mathcal{L}_{cls}$ and $\mathcal{L}_{p}$ enable the model to be supervised by labeled source and pseudo-labeled target samples. $\mathcal{L}_{adv}$ helps to learn domain-invariant features and minimize the distribution discrepancy. $\mathcal{L}_{ac}$ and $\mathcal{L}_{as}$ enhance the model robustness and reduce visual confusion on the target domain, thus improving target performance. The overall pipeline of UARN is illustrated in Algorithm \ref{al1}.


\begin{algorithm}
  \SetAlgoLined
  \SetKwInOut{Input}{Input}\SetKwInOut{Output}{Output}
  \Input{Labeled source data $\mathcal{D}^s=\{{x^{s}_{i}},{y^{s}_{i}}\}^{N_s}_{i=1}$, and unlabeled target data $\mathcal{D}^t=\{{x^{t}_{i}}\}^{N_t}_{i=1}$.}
  \Output{The trained recognition model $f$.}
  \BlankLine
  Initialize $E$ with the pretrained ImageNet model\;
  \While {network not converge}{
  $\left \{ x^s_i, y^s_i \right \}_{i=1}^B \leftarrow$ SampleMiniBatch$\left (  \mathcal{D}^s, B\right )$\;
  $\left \{ x^t_i, T(x^t_i) \right \}_{i=1}^B \leftarrow$ SampleMiniBatch$\left (  \mathcal{D}^t, B\right )$\;
  Compute $\mathcal{L}_{cls}$ by Eq. (\ref{source})\;
  Compute $\mathcal{L}_{adv}$ by Eq. (\ref{adversarial})\;
  \For{$i=1$ \KwTo $B$}{
  Generate $\hat{y}_i^t$ based on $f\left ( x_i^t  \right )$ by Eq. (\ref{pseudo})\;
  Compute $\mathcal{A}^{i,t}=g(x_i^t)$ and $\overline{\mathcal{A}^{i,t}} =g\left(T\left(x_i^t\right)\right)$\;
  }
  Compute $\mathcal{L}_{p}$ by Eq. (\ref{target})\;
  Compute $\mathcal{L}_{ac}$ by Eq. (\ref{consistency})\;
  Compute $\mathcal{L}_{as}$ by Eq. (\ref{separable})\;
  Optimize $E$ and $G$ by minimizing $\mathcal{L}_{cls}+\mathcal{L}_{p}$
  \ \ \ \ \ $+ \mathcal{L}_{adv}+\mu \mathcal{L}_{ac}+\lambda \mathcal{L}_{as}$\;
  Optimize $D$ by maximizing $\mathcal{L}_{adv}$\;
  }
  \caption{Pseudo code of UARN.}\label{al1}
\end{algorithm}

\subsection{Discussion}

\textbf{Comparison with BSP \cite{chen2019transferability}.} Learning discriminative features for the target domain is a hot topic in UDA since simply aligning domains cannot reach the optimal target performance. To this end, BSP \cite{chen2019transferability} proposed to penalize the largest singular values. Different from it, our UARN aims to learn discriminative features from a new perspective, which takes interpretability into consideration and enforces the separability of attention heatmaps.

\textbf{Comparison with STSN \cite{9757826}.}  To our knowledge, there is only one work focusing on UDA of oracle character recognition. STSN \cite{9757826} utilized GAN \cite{goodfellow2014generative} to transform handprinted oracle characters into scanned ones. Although the transformed scanned images can improve the target performance, the optimization of GAN suffers from instability and the supervision with cross-entropy loss only obtains the limited improvement. Contrarily, we leverage attention regularizations to achieve robustness and discrimination, leading to superior performance.

\textbf{Comparison with VAC \cite{guo2019visual}.} Our approach is most related to VAC, which regulars attention consistency under spatial transformations. However, our objective and algorithm are different from those of VAC. First, VAC addresses the problem of multi-label image classification, whereas we focus on oracle character recognition. Second, we further constrain the attention separability to achieve discrimination. Third, traditional attention regularizations are performed under the supervised learning setting, while UARN modifies and incorporates them into UDA framework by employing pseudo-class. Moreover, prediction confidence and entropy-aware weight are introduced to reduce the negative effect of hard-to-transfer examples.

\section{Experiments}

In this section, we evaluate the proposed method on oracle character recognition and digit classification  with state-of-the-art (SOTA) domain adaptation methods. In addition, we conduct ablation study, parameter sensitivity analysis and visualization to examine the contribution of our design to performance improvement.

\subsection{Datasets}

\begin{figure}
\centering
\subfigure[Handprinted oracle characters]{
\label{hand_ex} 
\includegraphics[height=3.1cm]{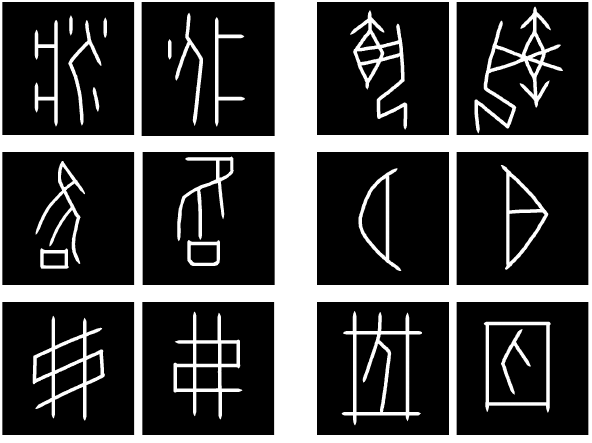}}
\hspace{0cm}
\subfigure[Scanned oracle characters]{
\label{scan_ex} 
\includegraphics[height=3.1cm]{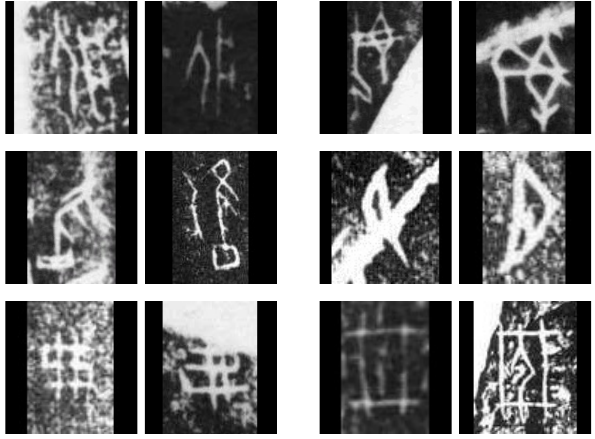}}
\caption{Examples of handprinted and scanned characters in Oracle-241.}
\label{Oracle241} 
\end{figure}

\textbf{Oracle-241} \cite{9757826}  is a benchmark dataset for domain adaptation of oracle character recognition, as shown in Fig. \ref{Oracle241}. It contains 80K images from 241 categories of oracle characters, shared by two significantly different domains. The domain of handprinted oracle characters contains labeled 10,861 samples for training and 3,730 samples for testing; while the domain of scanned oracle characters consists of unlabeled 50,168 samples for training and 13,806 samples for testing. Following \cite{9757826}, we transfer knowledge from handprinted oracle data to scanned oracle characters.

\textbf{MNIST-USPS-SVHN \cite{lecun1998gradient,netzer2011reading,denker1989neural}} are digits classification datasets containing 10 classes of digits. MNIST (M) and USPS (U) are handwritten digit datasets with grey-scale images. SVHN (S) consists of colored images obtained by detecting house numbers from Google Street View images. We follow previous work \cite{hoffman2018cycada} to construct three transfer tasks: M$\rightarrow$U, U$\rightarrow$M and S$\rightarrow$M.

\subsection{Implementation detail}

\textbf{Network architecture.} We adopt a ResNet-18 \cite{he2016deep} pre-trained on ImageNet \cite{russakovsky2015imagenet} as the feature extractor in the experiments for oracle character recognition. For the experiments on digits datasets, we use the sample LeNet architecture following previous works \cite{hoffman2018cycada,long2018conditional}. The domain discriminator consists of two layers with ReLU and Dropout (0.5) in all the layers, which shares the same architecture with DANN \cite{Ganin2015Unsupervised}.

\begin{table}
\renewcommand\arraystretch{1.1}
\caption{Source and target accuracies (mean$\pm$std\%) on Oracle-241 dataset. The best accuracy is
indicated in \textbf{\textcolor{red}{bold red}} and the second best accuracy is indicated in \textcolor{blue}{\underline{underlined blue}}.}
 \label{oracle-241}
	\begin{center}
    \small
    \setlength{\tabcolsep}{3mm}{
	\begin{tabular}{c|l|ccc}
		\toprule
           \multicolumn{2}{c|}{Methods} & Source: & Target: \\
         \multicolumn{2}{c|}{ } & Handprint & Scan  \\ \hline \hline
         Source-  & ResNet \cite{huang2019obc306} & \textcolor{blue}{\underline{94.9$\pm $0.1}}  &  2.1$\pm $0.6   \\
         only &NN-DML \cite{zhang2019oracle} & 94.5$\pm $0.4  &  8.4$\pm $1.0   \\\hline \hline
         \multirow{14}{*}{UDA}& CORAL \cite{sun2016deep} & 89.5$\pm $0.6 &  18.4$\pm $1.3 \\
         & DDC \cite{tzeng2014deep} & 90.8$\pm $1.5  &   25.6$\pm $1.9   \\
         & DAN \cite{long2018transferable} & 90.2$\pm $1.5  &   28.9$\pm $1.6   \\
         & ASSDA \cite{zhang2021robust} &85.8$\pm $0.1 &32.6$\pm $0.2 \\
         & DANN \cite{Ganin2015Unsupervised} & 87.1$\pm $1.7 &  32.7$\pm $1.5     \\
         & GVB \cite{cui2020gradually} & 92.8$\pm $0.4  & 36.8$\pm $1.1  \\
         & CDAN \cite{long2018conditional} & 85.3$\pm $3.3 & 37.9$\pm $2.0  \\
         & MSTN \cite{xie2018learning} & 91.0$\pm $1.5 & 38.3$\pm $1.2 \\
         & TransPar \cite{9807644} & 93.1$\pm $0.4 & 39.8$\pm $1.1 \\
         & FixBi \cite{na2021fixbi} & 90.1$\pm $1.6 & 40.2$\pm $0.1 \\
         & PRONOUN \cite{hu2021adversarial} & 92.4$\pm $0.3 & 40.3$\pm $1.8  \\
         & BSP \cite{chen2019transferability} &87.7$\pm $0.7 & 43.7$\pm $0.4  \\
         & STSN \cite{9757826}  & \textbf{\textcolor{red}{95.0$\pm $0.2}} &  \textcolor{blue}{\underline{47.1$\pm$0.8}}   \\ \cline{2-4}
         & \cellcolor{blue!5}\textbf{UARN (ours)}  & \cellcolor{blue!5}92.0$\pm $1.1 & \cellcolor{blue!5}\textbf{\textcolor{red}{55.6$\pm$0.9}}   \\
         \bottomrule
         \end{tabular}}
    \end{center}
\end{table}

\textbf{Experimental setup.} The experiments are implemented in Python on a desktop with one Tesla T4 GPU and Intel Xeon Gold 5218 CPU of 2.3GHz. We follow the standard protocols for UDA as \cite{Ganin2015Unsupervised,Long2015Learning}.  The average classification accuracy and the standard deviation of each adaptation task are reported on three random experiments.

For oracle character recognition, we resize the images to 224$\times $224, and pre-process them by random horizontal ﬂip and random erasing. Specifically, we randomly flip the images with 0.5 probability. The minimum and maximum area of the erased rectangle, i.e., $sl$ and $sh$, are 0.02 and 0.4, and the aspect ratio of the erased area, i.e., $r1$, is 0.3. We employ the mini-batch stochastic gradient descent (SGD) with momentum of 0.9. The model is trained for 100,000 iterations with the batch size of 36. We follow \cite{9757826} to employ the annealing strategy of learning rate. The initial learning rate $\eta_0$ is 0.001 and is adjusted using $\eta=\eta_0\left ( \frac{1-T}{T_{max}} \right )^{0.9}$, where $T$ and $T_{max}$ are the current and total iteration. The trade-off parameters $\mu$ and $\lambda$ are respectively set to 0.1 and 0.2, and the threshold $\tau$ is 0.85.

For digit classification, we compute the attention maps via grad-CAM \cite{selvaraju2017grad} since LeNet contains no GAP layers. Considering digits are sensitive to flipping, we randomly rotate target images with [$-10^\circ$,$10^\circ$] and reduce the distance between the attention map of the rotated image and the rotated attention map of the original image. The images are resized to 28$\times$28 in the U$\rightarrow$M and M$\rightarrow$U tasks, and 32$\times$32 in the S$\rightarrow$M task. We employ the mini-batch stochastic gradient descent (SGD) with momentum of 0.9. The model is trained for 40 epochs with the batch size of 64. The learning rate is set to 0.01 in the U$\rightarrow$M and M$\rightarrow$U tasks, and set to 0.003 in the S$\rightarrow$M task. The trade-off parameters $\mu$ and $\lambda$ are respectively set to 0.3 and 0.005, and the threshold $\tau$ is 0.95.

\subsection{Comparison with state-of-the-arts}

\textbf{Results on Oracle-241.} Table \ref{oracle-241} reports the performance comparison between our proposed model and other competing SOTA methods on Oracle-241 dataset. We aim to transfer knowledge from handprinted data to scanned characters and thus improve the performance on scanned data. We have the following conclusions summarized from Table \ref{oracle-241}. (1) Source-only methods, which train the models on handprinted oracle data without adaptation, only obtain the accuracies of less than 10\% on real-world scanned oracle characters, demonstrating the existence of cross-domain discrepancy. (2) Existing UDA methods substantially outperform source-only methods. This validates that explicitly reducing the cross-domain discrepancy can learn more transferable features. DANN \cite{Ganin2015Unsupervised}, ASSDA \cite{zhang2021robust} and CDAN \cite{long2018conditional}] employ adversarial training to learn domain-invariant features, while MSTN \cite{xie2018learning} utilizes pseudo-labeling to incorporate the semantic information into target training. However, these approaches fall short of ensuring sufficient robustness and discriminative capabilities for the model. (3) STSN \cite{9757826} is the first work focusing on UDA of oracle character recognition, which achieves the second-best performance on the target domain through joint disentanglement, transformation and adaptation. (4) Our UARN achieves the SOTA adaptation performance on Oracle-241, significantly surpassing CDAN and MSTN by 17.7\% and 17.3\%, respectively, in terms of target accuracy. This result underscores the importance of learning robust and discriminative features on the target domain. Compared with the best UDA competitors, i.e., STSN, UARN increases the target accuracy from 47.1\% to 55.6\%. The adaptation performance of STSN is largely determined by the quality of generation, whereas our method is simpler and more efficient. Although the source performance is slightly decreased since UARN emphasizes more on target domain compared with source-only methods, it remains to be 92.0\% and is superior to BSP \cite{chen2019transferability}.

\begin{table}
\renewcommand\arraystretch{1.1}
\caption{Target accuracies (mean$\pm$ std\%) on three transfer tasks of digit datasets. The best accuracy is
indicated in \textbf{\textcolor{red}{bold red}} and the second best accuracy is indicated in \textcolor{blue}{\underline{underlined blue}}.}
\label{MNIST-USPS-SVHN}
	\begin{center}
    \small
    \setlength{\tabcolsep}{2.3mm}{
	\begin{tabular}{l|ccc|c}
		\toprule
         Methods &   U$\rightarrow $M  & M$\rightarrow $U & S$\rightarrow $M & \cellcolor{gray!7} Avg \\ \hline \hline
         Source-only \cite{he2016deep} & 69.6$\pm$3.8 & 82.2$\pm$0.8 & 67.1$\pm$0.6 & \cellcolor{gray!7} 73.0 \\
         DANN \cite{Ganin2015Unsupervised} & - & 77.1$\pm$1.8 & 73.6& \cellcolor{gray!7} - \\
         DRCN \cite{ghifary2016deep} & 73.7$\pm$0.1 & 91.8$\pm$0.1 & 82.0$\pm$0.2 & \cellcolor{gray!7} 82.5 \\
         ADDA \cite{Tzeng2017Adversarial} & 90.1$\pm$0.8 & 89.4$\pm$0.2 & 76.0$\pm$1.8 & \cellcolor{gray!7} 85.2 \\
         DAA \cite{jia2019domain} &  92.8$\pm$1.1 & 90.3$\pm$0.2 & 78.3$\pm$0.5 & \cellcolor{gray!7} 87.1 \\
         LEL \cite{luo2017label} & - & - & 81.0$\pm$0.3& \cellcolor{gray!7} - \\
         DSN \cite{bousmalis2016domain} & - & - & 82.7& \cellcolor{gray!7} - \\
         DTN \cite{taigman2016unsupervised} & - & - & 84.4& \cellcolor{gray!7} - \\
         ARTN \cite{cai2019unsupervised}& - & - & 85.8 &  \cellcolor{gray!7} - \\
         AsmTri \cite{saito2017asymmetric} &   - & - & 86.0& \cellcolor{gray!7} - \\
         CoGAN \cite{liu2016coupled}  & 89.1$\pm$0.8 & - & 91.2$\pm$0.8& \cellcolor{gray!7} - \\
         GTA \cite{sankaranarayanan2018generate} & 90.8$\pm$1.3 & 92.8$\pm$0.9 & 92.4$\pm$0.9 & \cellcolor{gray!7} 92.0 \\
         MSTN \cite{xie2018learning} & - & 92.9$\pm$1.1 & 91.7$\pm$1.5& \cellcolor{gray!7} - \\
         PixelDA \cite{bousmalis2017unsupervised} & & \textcolor{red}{\textbf{95.9}} & & \cellcolor{gray!7} -\\
         SRDA \cite{cai2021learning} & 96.0 & 93.3 & 89.5 & \cellcolor{gray!7} 92.9 \\
         TPN  \cite{pan2019transferrable} & 94.1 &  92.1 & \textcolor{blue}{\underline{93.0}} & \cellcolor{gray!7} 93.1 \\
         UNIT \cite{liu2017unsupervised} & 93.6 & \textcolor{red}{\textbf{95.9}} & 90.5 & \cellcolor{gray!7} 93.4\\
         DSAN \cite{zhu2020deep} & \textcolor{blue}{\underline{96.9$\pm$0.2}} & 95.3$\pm$0.1 & 90.1$\pm$0.4 & \cellcolor{gray!7} 94.1 \\
         CyCADA \cite{hoffman2018cycada} & 96.5$\pm$0.1 & \textcolor{blue}{\underline{95.6$\pm$0.2}} & 90.4$\pm$0.4 & \cellcolor{gray!7} 94.2 \\
         STSN \cite{9757826} & 96.7$\pm$0.1 &  94.4$\pm$0.3 &  92.2$\pm$0.1 & \cellcolor{gray!7} \textcolor{blue}{\underline{94.4}} \\
         \hline \rowcolor{blue!5}
         \textbf{UARN (ours)} &  \textcolor{red}{\textbf{97.6$\pm$0.3}} &  94.8$\pm$0.1 &  \textcolor{red}{\textbf{93.3$\pm$0.7}} &\cellcolor{gray!7} \textcolor{red}{\textbf{95.2}}  \\
         \bottomrule
	\end{tabular}}
    \end{center}
\end{table}

\textbf{Results on MNIST-USPS-SVHN.} Table \ref{MNIST-USPS-SVHN} reports the target accuracy on digit datasets to prove that UARN has the potential to generalize to other character recognition task. It is important to note that we enforce attention consistency under the rotation transformation since digits are sensitive to flipping. We have some essential observations from the performance in Table \ref{MNIST-USPS-SVHN}.  Our model obtains 97.6\%, 94.8\% and 93.3\% on the tasks of U$\rightarrow $M, M$\rightarrow $U and S$\rightarrow $M, respectively. Compared with existing advanced methods, our UARN performs better and achieves higher average accuracy than CyCADA \cite{hoffman2018cycada} and STSN \cite{9757826} by 1.0\% and 0.8\%, especially in the extremely hard task S$\rightarrow $M. Existing UDA methods ignore visual perceptual plausibility when adapting, and thus result in sub-optimal performance on the target domain; while our UARN enforces the consistency and discriminability of attention heatmaps to improve the model robustness and reduce visual confusion.

\subsection{Ablation study}

\begin{table}
\renewcommand\arraystretch{1.1}
\small
\caption{Ablation investigations of our model on Oracle-241 dataset. ACC means the accuracy on scanned data.}
\label{ablation}
	\begin{center}
    \setlength{\tabcolsep}{3mm}{
	\begin{tabular}{ccccc|cc}
		\toprule
         $\mathcal{L}_{cls}$  & $\mathcal{L}_{adv}$  & $\mathcal{L}_{p}$ & $\mathcal{L}_{ac}$ & $\mathcal{L}_{as}$  & ACC & $\Delta$(\%) \\ \hline \hline
         \ding{51}& \textcolor{red}{\ding{55}} & \textcolor{red}{\ding{55}} & \textcolor{red}{\ding{55}}  & \textcolor{red}{\ding{55}} &2.1 & -  \\
         \ding{51}& \ding{51} & \textcolor{red}{\ding{55}} & \textcolor{red}{\ding{55}}  & \textcolor{red}{\ding{55}}& 44.1 & $\uparrow$42.0  \\
         \ding{51}&  \ding{51} & \ding{51} & \textcolor{red}{\ding{55}} & \textcolor{red}{\ding{55}} & 51.0 & $\uparrow$48.9   \\
         \ding{51}& \ding{51}& \ding{51} & \ding{51} & \textcolor{red}{\ding{55}} & 54.7 & $\uparrow$52.6  \\
         \ding{51}& \ding{51}& \ding{51} & \ding{51} & \ding{51} & \textbf{55.6} & $\uparrow$53.5  \\
         \bottomrule
         \end{tabular}}
    \end{center}
\end{table}

\begin{table}
\renewcommand\arraystretch{1.1}
\small
\caption{Ablation investigations of our model on the S$\rightarrow $M task of digit datasets. ACC means the accuracy on the target domain.}
\label{ablation_digit}
	\begin{center}
    \setlength{\tabcolsep}{3mm}{
	\begin{tabular}{ccccc|cc}
		\toprule
         $\mathcal{L}_{cls}$  & $\mathcal{L}_{adv}$  & $\mathcal{L}_{p}$ & $\mathcal{L}_{ac}$ & $\mathcal{L}_{as}$  & ACC & $\Delta$(\%) \\ \hline \hline
         \ding{51}& \textcolor{red}{\ding{55}} & \textcolor{red}{\ding{55}} & \textcolor{red}{\ding{55}}  & \textcolor{red}{\ding{55}} &65.1 & -  \\
         \ding{51}& \ding{51} & \textcolor{red}{\ding{55}} & \textcolor{red}{\ding{55}}  & \textcolor{red}{\ding{55}}& 67.7 & $\uparrow$2.6  \\
         \ding{51}&  \ding{51} & \ding{51} & \textcolor{red}{\ding{55}} & \textcolor{red}{\ding{55}} & 85.5 & $\uparrow$20.4   \\
         \ding{51}& \ding{51}& \ding{51} & \ding{51} & \textcolor{red}{\ding{55}} & 91.8 & $\uparrow$26.7  \\
         \ding{51}& \ding{51}& \ding{51} & \ding{51} & \ding{51} & \textbf{93.3} & $\uparrow$28.2  \\
         \bottomrule
         \end{tabular}}
    \end{center}
\end{table}

\textbf{Effectiveness of each component.} We conduct ablation experiments on Oracle-241 dataset to investigate the effects of different components in our UARN, as shown in Table \ref{ablation}. We denote the method only using $\mathcal{L}_{cls}$ as \emph{BASE}, and denote \emph{BASE}+$\mathcal{L}_{adv}$+$\mathcal{L}_{p}$ as \emph{BASE-adapt}. (1) It can be found that \emph{BASE-adapt} significantly outperforms \emph{BASE} by introducing adversarial learning $\mathcal{L}_{adv}$ and pseudo labeling $\mathcal{L}_{p}$. $\mathcal{L}_{adv}$ helps to minimize the distribution discrepancy across domains, and $\mathcal{L}_{p}$ enables the model optimization on the target domain via self-training. (2) When adding the attention consistency loss $\mathcal{L}_{ac}$ to \emph{BASE-adapt}, our UARN further achieves gains of 3.7\% in terms of target accuracy. It illustrates the effectiveness and importance of $\mathcal{L}_{ac}$ in our model to enhance the model robustness to flipping and prevent the model from attending to irrelevant regions of the flipped images. We note that data augmentation, i.e., random horizontal flip, is applied in UARN and its variants including \emph{BASE-adapt}. However, \emph{BASE-adapt} cannot achieve competitive results compared with our method. (3) The performance of our UARN undergoes a decrease of 0.9\% when we remove the attention discriminative loss $\mathcal{L}_{as}$, which justifies the effectiveness of this module to make the attention maps separable and tell the confusing classes apart.

In Table \ref{ablation_digit}, similar observations can be obtained from the ablation study on digit datasets. We observe that adding $\mathcal{L}_{ac}$ boosts the performances by 6.3\%, and the model’s performance drops from 93.3\% to 91.8\% when $\mathcal{L}_{as}$ is removed from our UARN. It also demonstrates that each part has a specific contribution.

\textbf{Spacial transformation.} Our proposed UARN constrains attention consistency under the flipping transformation. To verify its effectiveness and superiority, we compare flipping with other spacial transformations on Oracle-241, i.e., rotation and scaling. For rotation, we randomly rotate the target image with [$-10^\circ$,$10^\circ$], and reduce the distance between the attention map of the rotated image and the rotated attention map of the original image. For scaling, the target image is downscaled from 224$\times$224 to 196$\times$196. Then, we generate the attention map with the size of 7$\times$7 for the original image, and one with the size of 6$\times$6 for the scaling image. Finally, we upscale both the attention maps to 42$\times$42, and minimize their divergence. The comparison results are shown in Table \ref{transform}. We denote UARN w/o $\mathcal{L}_{ac}+\mathcal{L}_{as}$ as \emph{BASE-adapt}. It can be observed that constraining attention consistency under flipping is more effective and achieves higher target accuracy compared with rotation and scaling.

\begin{table}
\renewcommand\arraystretch{1.1}
\small
\caption{Comparisons with other transformation methods on Oracle-241. ACC means the accuracy on scanned data.}
\label{transform}
	\begin{center}
    \setlength{\tabcolsep}{5.5mm}{
	\begin{tabular}{l|cc}
		\toprule
         Transform &  ACC & $\Delta$(\%) \\ \hline \hline
         \emph{BASE-adapt} & 51.0 & - \\
         UARN w/ rotation &   52.4 & $\uparrow$ 1.4   \\
         UARN w/ scaling &  52.9 & $\uparrow$ 1.9  \\
         UARN w/ flipping (ours) &  \textbf{55.6} & $\uparrow$ 4.6  \\
         \bottomrule
         \end{tabular}}
    \end{center}
\end{table}

\textbf{Consistency regularization.} Here we study the effects of different consistency regularizations on adaptation performance. Inspired by \cite{sohn2020fixmatch}, UARN w/ PC enforces the consistency of model predictions under the flipping transformation. Specifically, it generates pseudo labels on the original images, and then trains the network to minimize cross entropy between the generated pseudo labels and the model’s outputs of the flipped images. As depicted in Table \ref{prediction consist}, our UARN (w/ AC) is superior to UARN w/ PC, and clearly improves the target performance from 53.5\% to 55.6\% on Oracle-241 and from 87.5\% to 93.3\% on the S$\rightarrow $M task of digit datasets. We believe the reason is as follows. Compared with the model prediction, attention heatmaps take advantage of more visual knowledge to precisely encode the models’ representation. Therefore, constraining attention consistency to minimize the divergence between the representation of $x^t_i$ and that of $T(x^t_i)$  is more meaningful and effective.

\begin{table}
\renewcommand\arraystretch{1.1}
\small
\caption{Comparison with prediction consistency on Oracle-241 and digit datasets. ACC means the accuracy on the target domain.}
\label{prediction consist}
	\begin{center}
    \setlength{\tabcolsep}{2.5mm}{
	\begin{tabular}{l|l|cc}
		\toprule
         Dataset & Method &  ACC & $\Delta$(\%) \\ \hline \hline
         \multirow{3}{*}{Oracle-241}& \emph{BASE-adapt} & 51.0 & - \\
         & UARN w/ PC &   53.5 & $\uparrow$ 2.5   \\
         & UARN w/ AC (ours) &  \textbf{55.6} & $\uparrow$ 4.6  \\ \hline
         \multirow{3}{*}{S$\rightarrow $M task}& \emph{BASE-adapt} & 85.5 & - \\
         & UARN w/ PC &   87.5 & $\uparrow$ 2.0   \\
         & UARN w/ AC (ours) &  \textbf{93.3} & $\uparrow$ 7.8  \\
         \bottomrule
         \end{tabular}}
    \end{center}
\end{table}

\begin{table}
\renewcommand\arraystretch{1.1}
\small
\caption{Ablation investigations of prediction confidence on Oracle-241 and digit datasets. ACC means the accuracy on the target domain.}
\label{confident}
	\begin{center}
    \setlength{\tabcolsep}{2.5mm}{
	\begin{tabular}{l|l|cc}
		\toprule
         Dataset & Method &  ACC & $\Delta$(\%) \\ \hline \hline
         \multirow{4}{*}{Oracle-241}& UARN (ours) &  \textbf{55.6} &  -  \\
         & UARN w/o AC$\tau$ &  54.0 & $\downarrow$ 1.6  \\
         & UARN w/o AS$\tau$ &  54.8 & $\downarrow$ 0.8  \\
         & UARN w/o $\varphi \left ( H(p_i) \right )$ &  55.1 & $\downarrow$ 0.5  \\ \hline
         \multirow{4}{*}{S$\rightarrow $M task}& UARN (ours) &  \textbf{93.3} &  -  \\
         & UARN w/o AC$\tau$ &  93.3 & $\downarrow$ 0.0  \\
         & UARN w/o AS$\tau$ &  88.7 & $\downarrow$ 4.6  \\
         & UARN w/o $\varphi \left ( H(p_i) \right )$ &  90.4 & $\downarrow$ 2.9  \\
         \bottomrule
         \end{tabular}}
    \end{center}
\end{table}

\textbf{Confidence of target samples.} Our attention regularizations are only performed on high-confident samples whose prediction confidences are higher than $\tau$. To investigate the effect of using high-confident samples, we compare UARN with two variants, i.e., UARN w/o AC$\tau$ and UARN w/o AS$\tau$. Specifically, UARN w/o AC$\tau$ performs attention consistency on all target samples and enforces attention separability on high-confident samples; while UARN w/o AS$\tau$ does the opposite. As shown in Table \ref{confident}, removing the constraint of high-confident samples in our attention regularizations results in a performance decrease of 0.8\%-1.6\% on Oracle-241. This is because low-confident samples may be falsely labeled and thus the quality of pseudo classes cannot be guaranteed in attention separability. Furthermore, the adapted model would fail to generate correct attention maps for low-confident samples. Applying regularizations on these noisy attention maps may lead to a negative influence. In Table \ref{confident}, we also verify the effectiveness of $\varphi \left ( H(p_i) \right )$ which reweights high-confident examples by their prediction confidences in attention separability. After removing $\varphi \left ( H(p_i) \right )$ and treating each sample equally, the performance of our model decreases by 0.5\% in terms of target accuracy on Oracle-241. It illustrates that the utilization of $\varphi \left ( H(p_i) \right )$ can further alleviate the negative influence caused by inaccurate pseudo classes. Moreover, we also conduct similar experiments on the S$\rightarrow $M task of digit datasets, and the results show that the model's performance declines when the related module is removed, demonstrating the importance of emphasizing high-confident samples.

\begin{figure*}
\centering
\subfigure[Sensitivity to $\mu$]{
\label{hand_ex} 
\includegraphics[height=2.8cm]{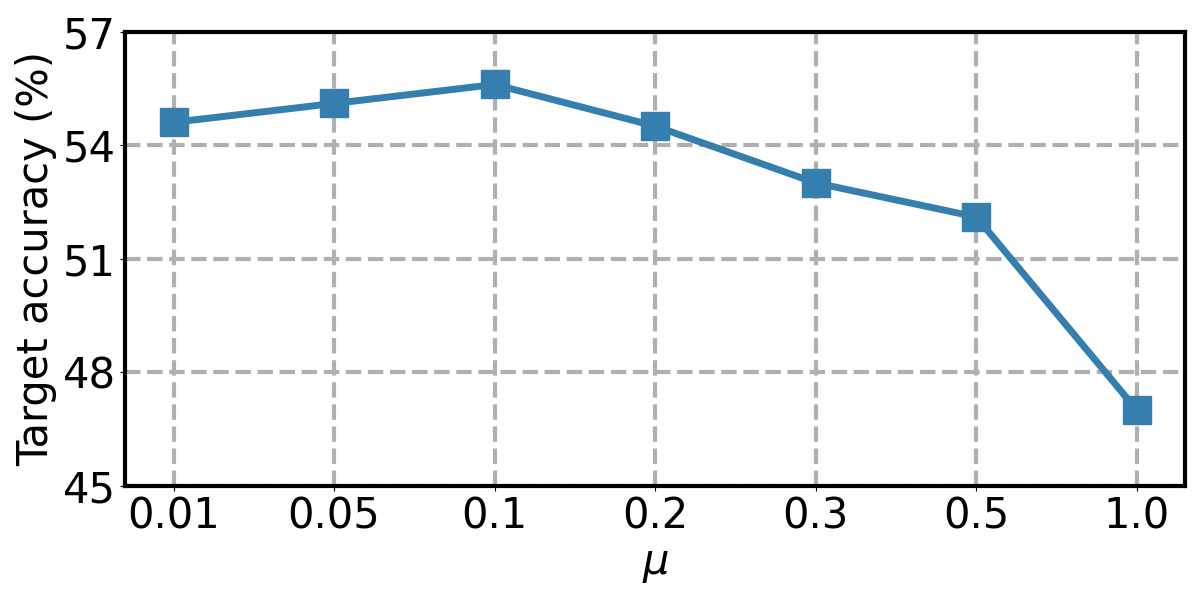}}
\hspace{1cm}
\subfigure[Sensitivity to $\lambda$]{
\label{scan_ex} 
\includegraphics[height=2.8cm]{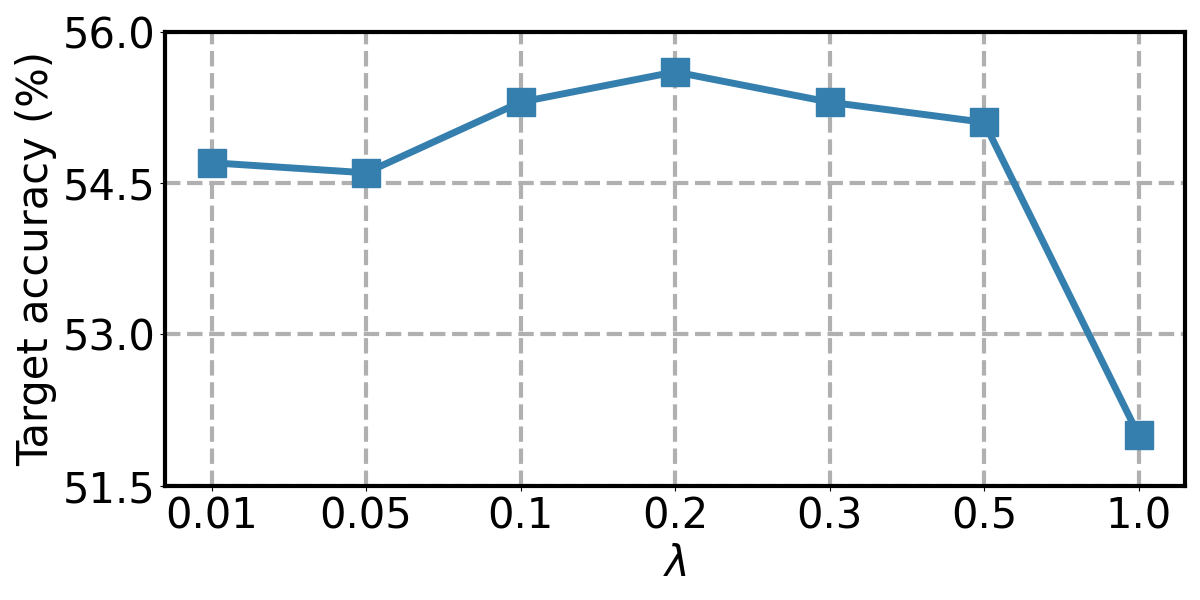}}
\caption{Parameter sensitivity investigations of $\mu$ and $\lambda$ in terms of target accuracy on Oracle-241 dataset.}
\label{sensitivity} 
\end{figure*}

\begin{figure*}
\centering
\subfigure[Sensitivity to $\alpha$]{
\label{hand_ex} 
\includegraphics[height=2.8cm]{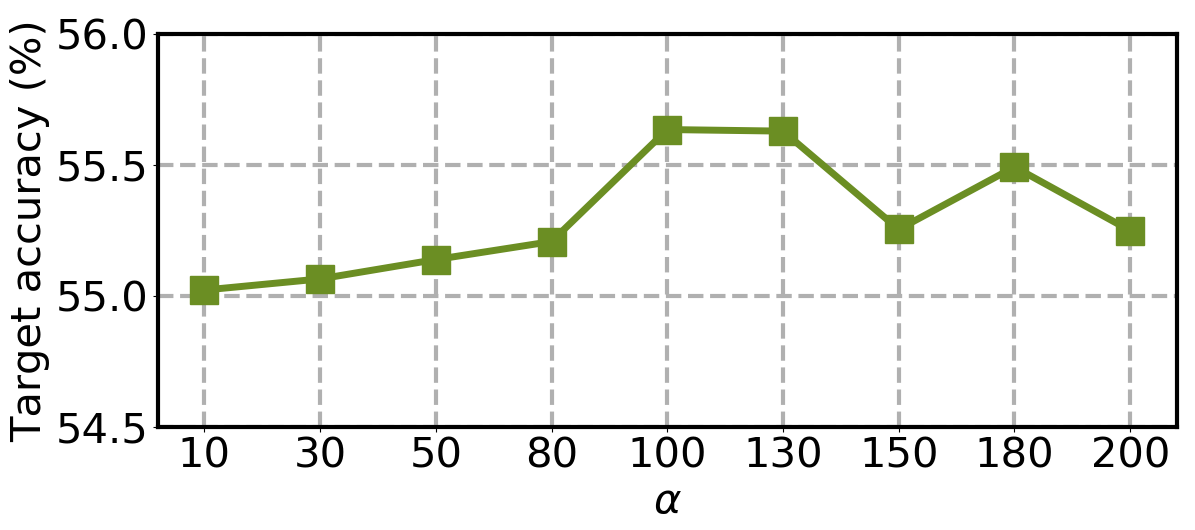}}
\hspace{1cm}
\subfigure[Sensitivity to $\beta$]{
\label{scan_ex} 
\includegraphics[height=2.8cm]{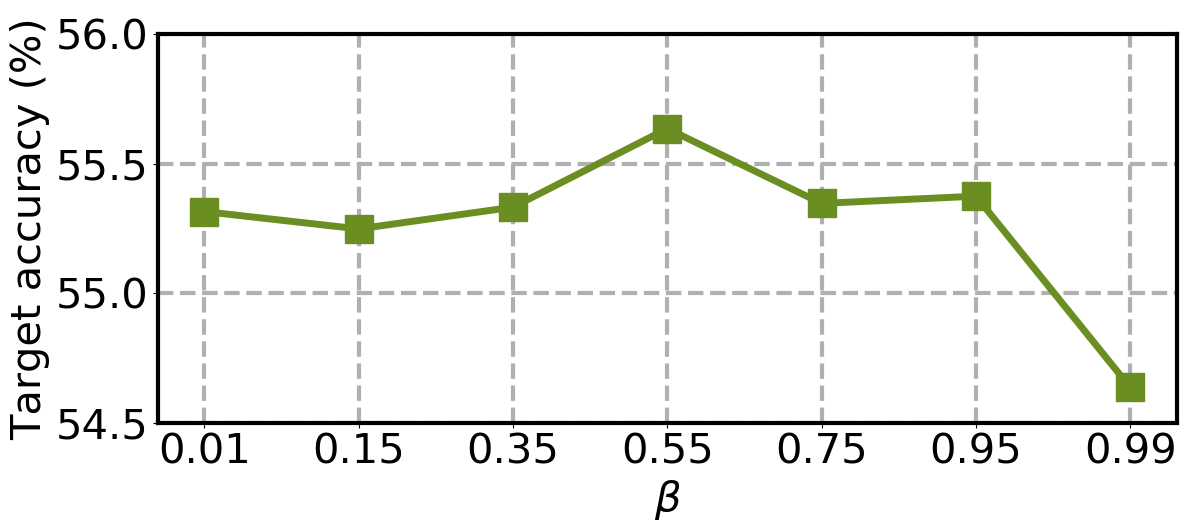}}
\caption{Parameter sensitivity investigations of $\alpha$ and $\beta$ in terms of target accuracy on Oracle-241 dataset.}
\label{sensi_alpha} 
\end{figure*}

\begin{figure*}
\centering
\subfigure[Mask rate]{
\label{thred_mask} 
\includegraphics[height=2.8cm]{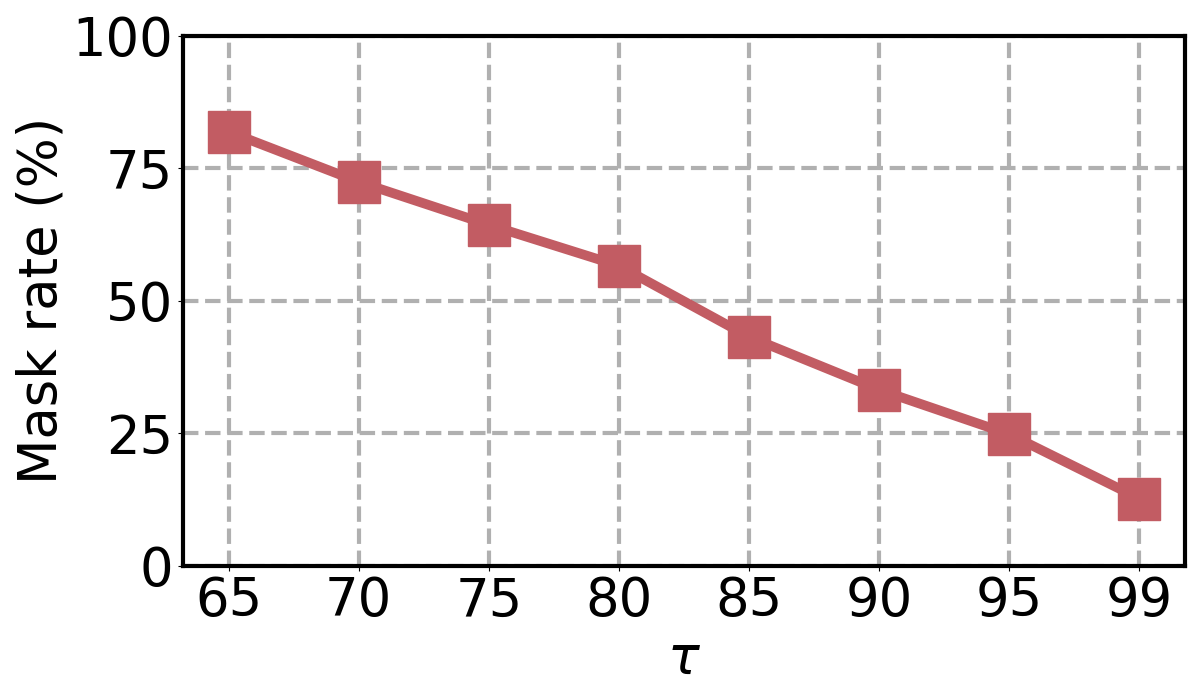}}
\hspace{0.5cm}
\subfigure[Purity]{
\label{thred_purity} 
\includegraphics[height=2.8cm]{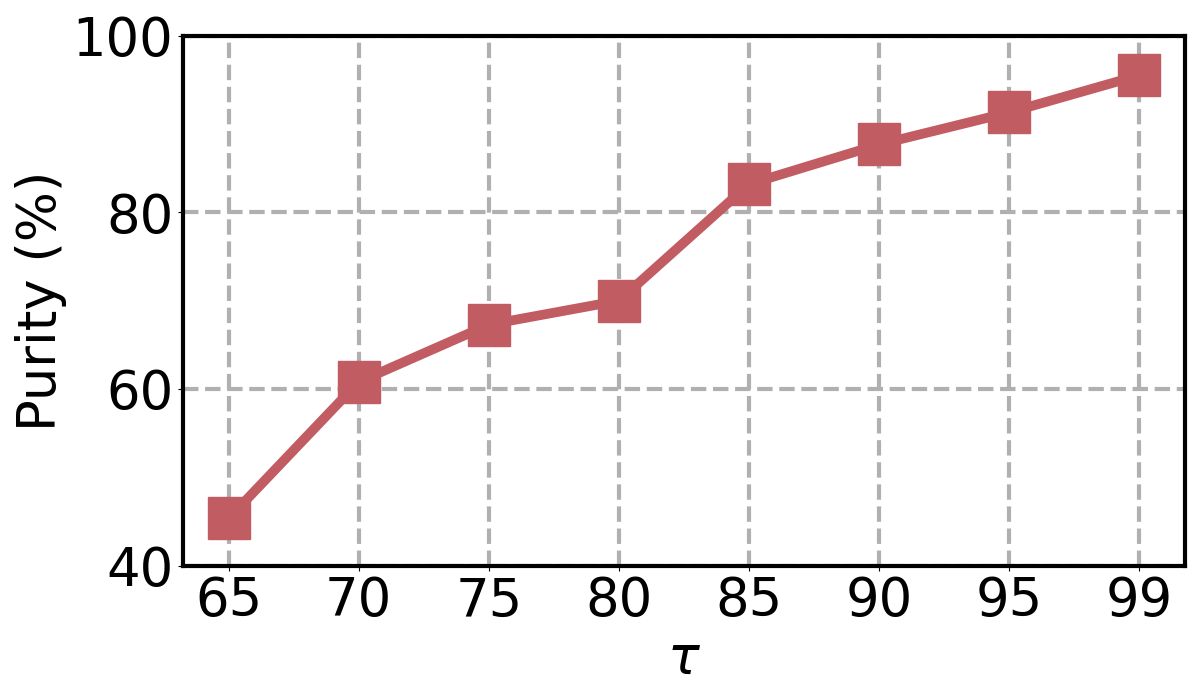}}
\hspace{0.5cm}
\subfigure[Target accuracy]{
\label{thred_acc} 
\includegraphics[height=2.8cm]{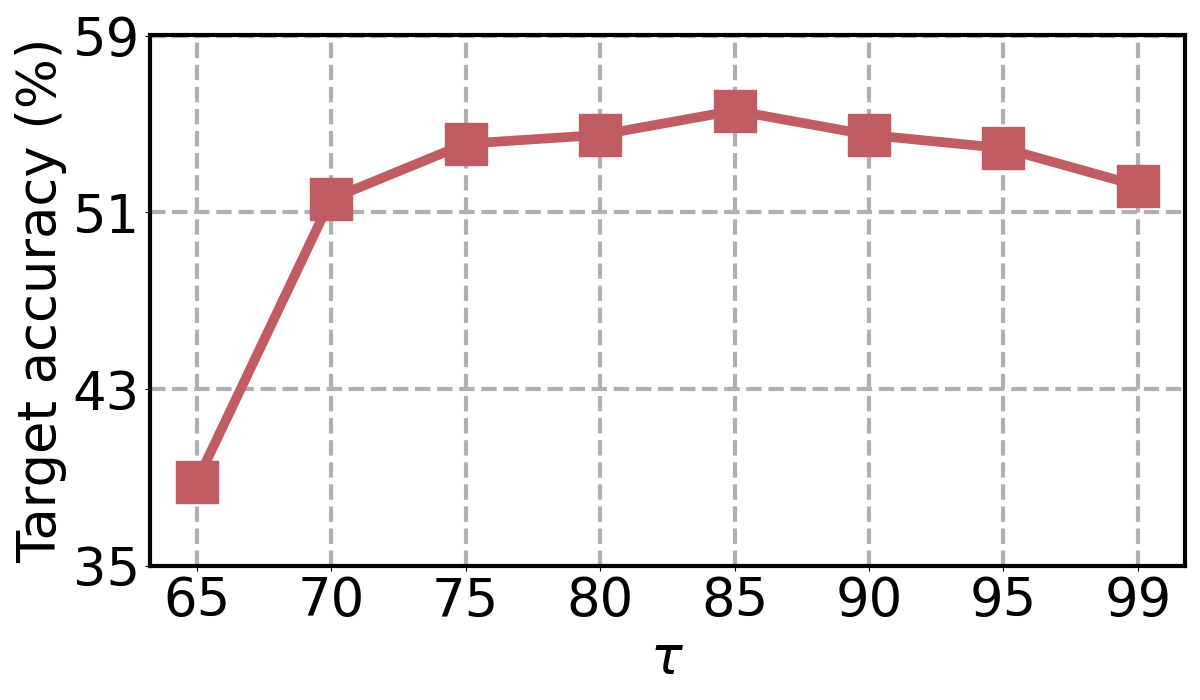}}
\caption{Parameter sensitivity investigations of $\tau$ in terms of (a) mask rate, (b) purity and (c) target accuracy on Oracle-241 dataset.}
\label{sensi_thred} 
\end{figure*}

\begin{figure*}
\centering
\subfigure[ResNet ]{
\label{visual1} 
\includegraphics[height=3.15cm]{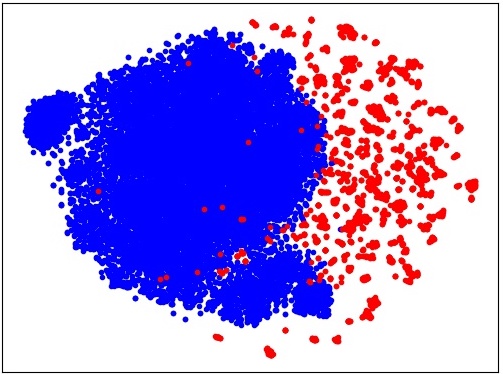}}
\hspace{0cm}
\subfigure[DANN \cite{Ganin2015Unsupervised}]{
\label{visual2} 
\includegraphics[height=3.15cm]{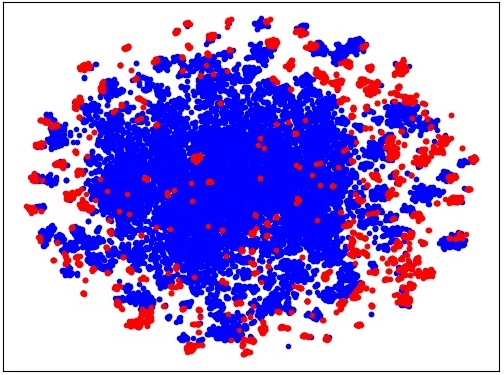}}
\hspace{0cm}
\subfigure[STSN \cite{9757826}]{
\label{visual2} 
\includegraphics[height=3.15cm]{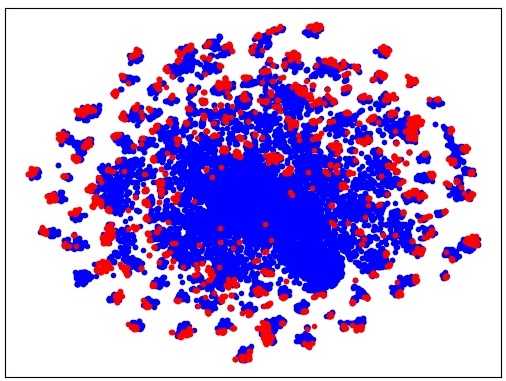}}
\hspace{0cm}
\subfigure[UARN (ours)]{
\label{visual4} 
\includegraphics[height=3.15cm]{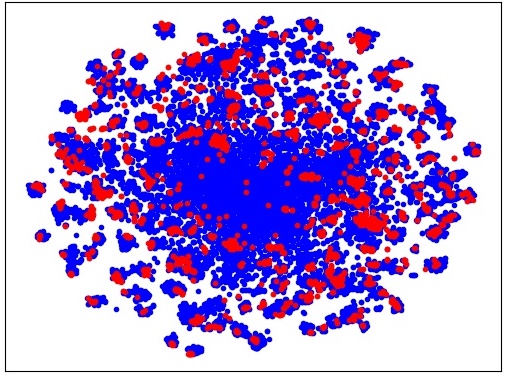}}
\caption{t-SNE \cite{maaten2008visualizing} embedding visualizations on Oracle-241. Colors denote different domains (red: handprinted data, blue: scanned data).}
\label{visual} 
\end{figure*}

\begin{figure*}
\centering
\subfigure[ResNet ]{
\label{visual_target_1} 
\includegraphics[height=3.15cm]{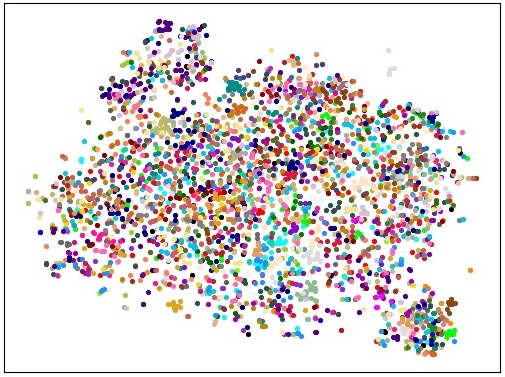}}
\hspace{0cm}
\subfigure[DANN \cite{Ganin2015Unsupervised}]{
\label{visual_target_2} 
\includegraphics[height=3.15cm]{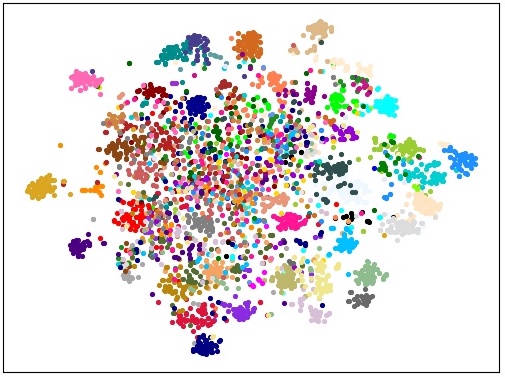}}
\hspace{0cm}
\subfigure[STSN \cite{9757826}]{
\label{visual_target_3} 
\includegraphics[height=3.15cm]{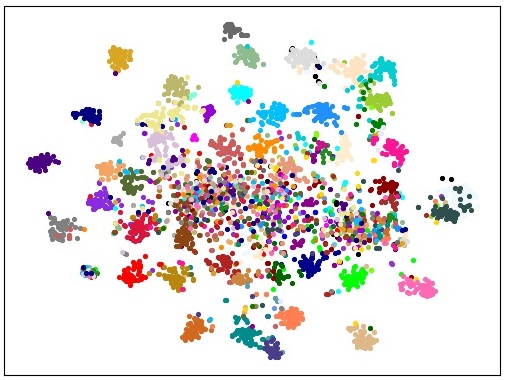}}
\hspace{0cm}
\subfigure[UARN (ours)]{
\label{visual_target_4} 
\includegraphics[height=3.15cm]{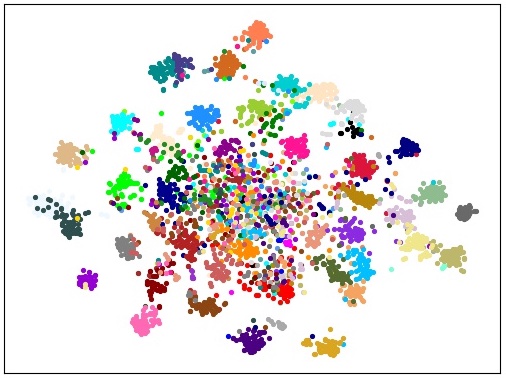}}
\caption{t-SNE \cite{maaten2008visualizing} embedding visualizations of different target classes on Oracle-241. Colors denote different classes.}
\label{visual_target} 
\end{figure*}

\begin{figure*}
\centering
\subfigure[ResNet ]{
\label{usps2mnist_1} 
\includegraphics[height=3.2cm]{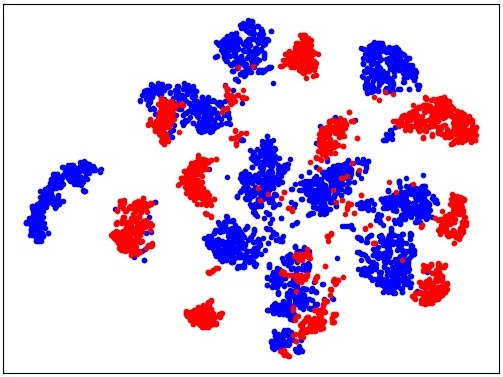}}
\hspace{0.8cm}
\subfigure[DANN \cite{Ganin2015Unsupervised}]{
\label{usps2mnist_2} 
\includegraphics[height=3.2cm]{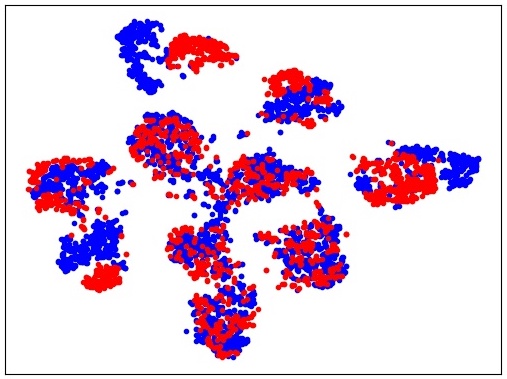}}
\hspace{0.8cm}
\subfigure[UARN (ours)]{
\label{usps2mnist_4} 
\includegraphics[height=3.2cm]{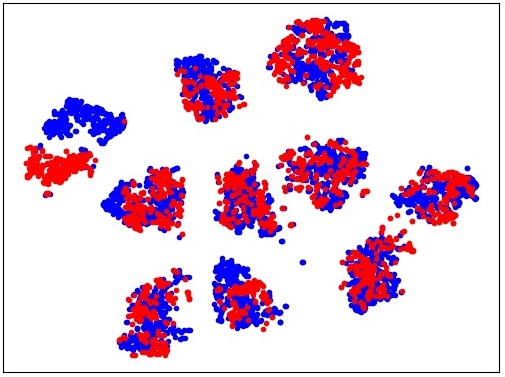}}
\caption{t-SNE \cite{maaten2008visualizing} embedding visualizations for the U$\rightarrow $M task on digit datasets. Colors denote different domains (red: USPS, blue: MNIST).}
\label{visual_usps2mnist} 
\end{figure*}

\begin{figure*}
\centering
\subfigure[ResNet ]{
\label{usps2mnist_1} 
\includegraphics[height=3.2cm]{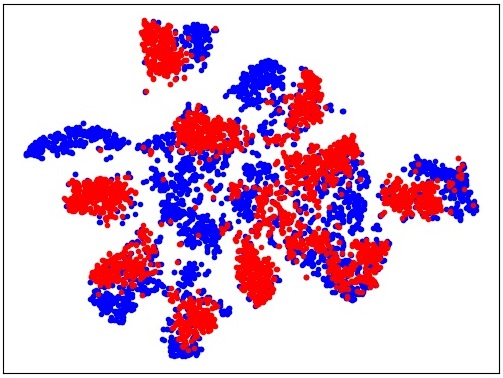}}
\hspace{0.8cm}
\subfigure[DANN \cite{Ganin2015Unsupervised}]{
\label{usps2mnist_2} 
\includegraphics[height=3.2cm]{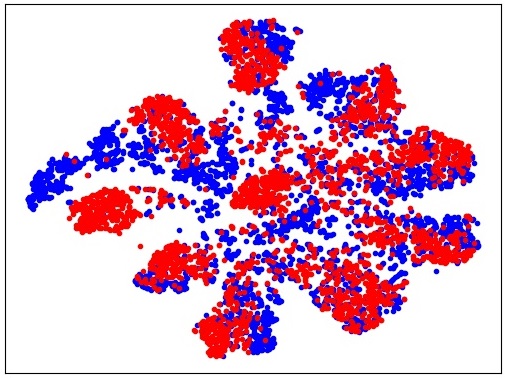}}
\hspace{0.8cm}
\subfigure[UARN (ours)]{
\label{usps2mnist_4} 
\includegraphics[height=3.2cm]{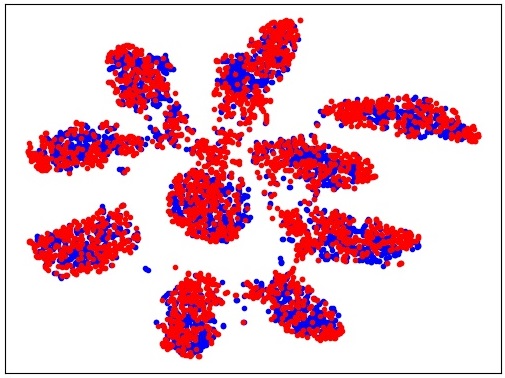}}
\caption{t-SNE \cite{maaten2008visualizing} embedding visualizations for the S$\rightarrow $M task on digit datasets. Colors denote different domains (red: SVHN, blue: MNIST).}
\label{visual_svhn2mnist} 
\end{figure*}

\begin{figure}
\centering
\subfigure[Source domain]{
\label{Convergence_a} 
\includegraphics[width=4.25cm]{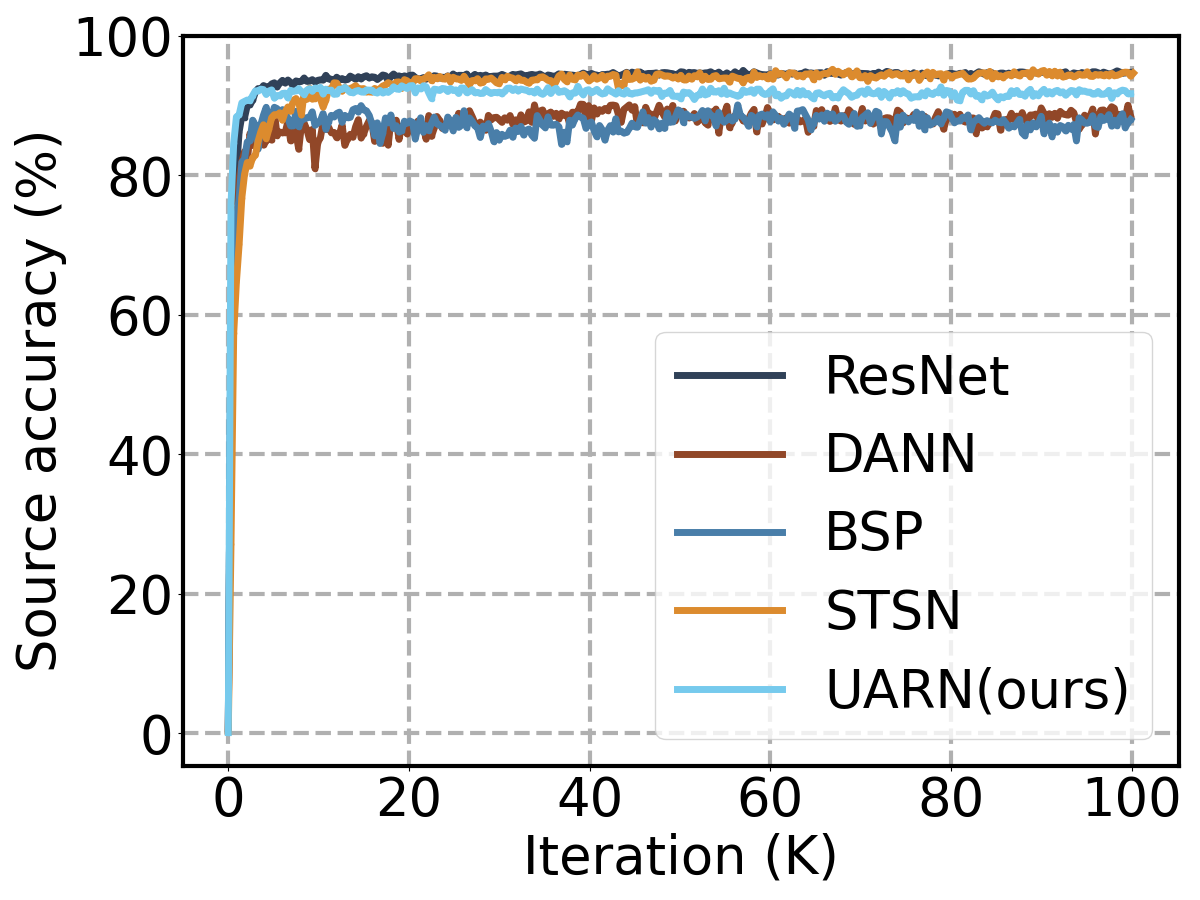}}
\hspace{-0.2cm}
\subfigure[Target domain]{
\label{Convergence_b} 
\includegraphics[width=4.25cm]{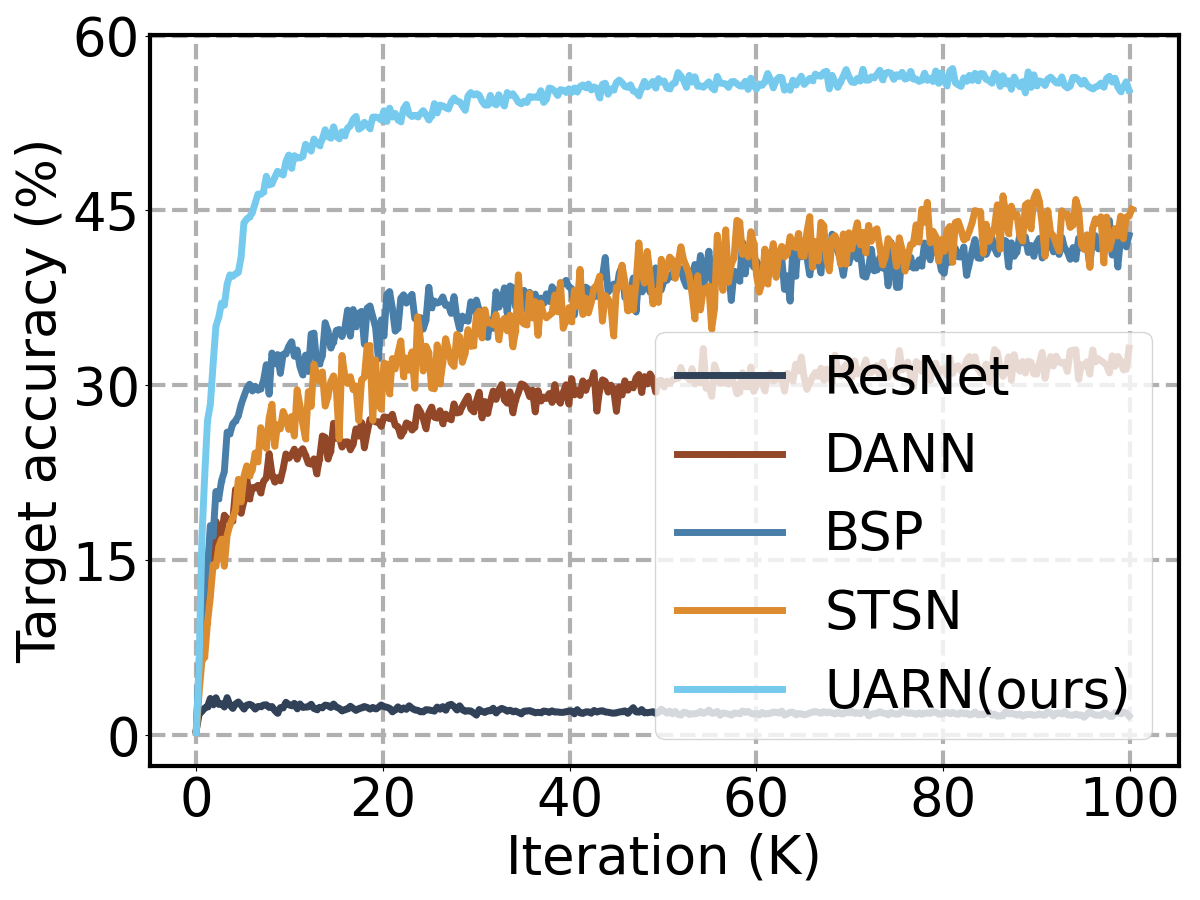}}
\caption{Convergence of ResNet, DANN \cite{Ganin2015Unsupervised}, BSP \cite{chen2019transferability}, STSN \cite{9757826} and our UARN on Oracle-241 dataset.}
\label{Convergence} 
\end{figure}

\subsection{Parameter sensitivity}

\textbf{Trade-off paramter $\mu$ and $\lambda$.} In UARN, $\mu$ and $\lambda$ are utilized to control the losses of $\mathcal{L}_{ac}$ and $\mathcal{L}_{as}$, respectively. To better understand their effects, we report the sensitivity of UARN to $\mu$ and $\lambda$ in Fig. \ref{sensitivity}. It can be observed that the target accuracy first increases and then decreases as $\mu$ and $\lambda$ vary. If $\mu$ and $\lambda$ are too small, the cross-entropy term will dominate the optimization and thus the resulting improvement in the interpretation consistency and discriminability will be marginal. Conversely, the extremely large values of $\mu$ and $\lambda$ would make the network overemphasize attention regularizations and weaken the effect of classification loss such that attention regularizations will be applied on noisy heatmaps resulting in lower accuracy. The best result is obtained at $\mu= 0.1$ and $\lambda= 0.2$.

\textbf{Trade-off paramter $\alpha$ and $\beta$.} We herein evaluate the sensitivity of the hyper-parameters involved in the attention discriminative loss, i.e., $\alpha$ and $\beta$ in Eq. (\ref{sepera8}). We vary $\alpha$ from 10 to 200 and $\beta$ from 0.01 to 0.99, with the results shown in Fig. \ref{sensi_alpha}. We observe that better results are generated when $\alpha=100$ and $\beta=0.55$. Inappropriate values will lead to a relatively weak constraint or result in the model excessively emphasizing high-response regions. However, the target accuracy is not so much sensitive to varying these hyper-parameters.

\textbf{Confidence threshold $\tau$.} We utilize the threshold $\tau$ to filter out low-confident samples in $\mathcal{L}_{p}$, $\mathcal{L}_{ac}$ and $\mathcal{L}_{as}$. In Fig. \ref{sensi_thred}, we study the sensitivity of UARN to $\tau$ in terms of mask rate, purity and target accuracy. Following \cite{sohn2020fixmatch}, we define mask rate (recall) and purity (precision) as,
\begin{equation}
\text{mask rate} = \frac{1}{N_t}\sum_{i=1}^{N_t}\mathbbm{1}\left ( \max p_i> \tau \right ),
\end{equation}
\begin{equation}
\text{purity} = \frac{\sum_{i=1}^{N_t}\mathbbm{1}\left ( \max p_i> \tau \right )\mathbbm{1}\left ( \hat{y}_i^t =y_i^t\right )}{\sum_{i=1}^{N_t}\mathbbm{1}\left ( \max p_i> \tau \right )},
\end{equation}
where $y_i^t$ is the ground-truth label of $x_i^t$. According to the definition, mask rate denotes the ratio of selected high-confident samples to all samples, and purity indicates the correctness of pseudo labels which are assigned to these high-confident samples. As depicted in Fig. \ref{sensi_thred}, mask rate decreases and purity increases with the increase of $\tau$. Since the samples whose confidences are lower than $\tau$ will be abandoned, using high threshold values will filter out lots of samples but ensure the quality of pseudo labels. 
When using small threshold values, most samples are remained and assigned pseudo labels. However, the learning process will be significantly impeded by noisy pseudo-labeled examples. Therefore, a proper value of $\tau$ is of vital importance for target performance to trade-off between the quality and quantity of high-confident samples and their pseudo labels. The best accuracy is obtained at $\tau= 0.85$, shown in Fig. \ref{thred_acc}.

\begin{figure*}
\centering
\includegraphics[width=18.2cm]{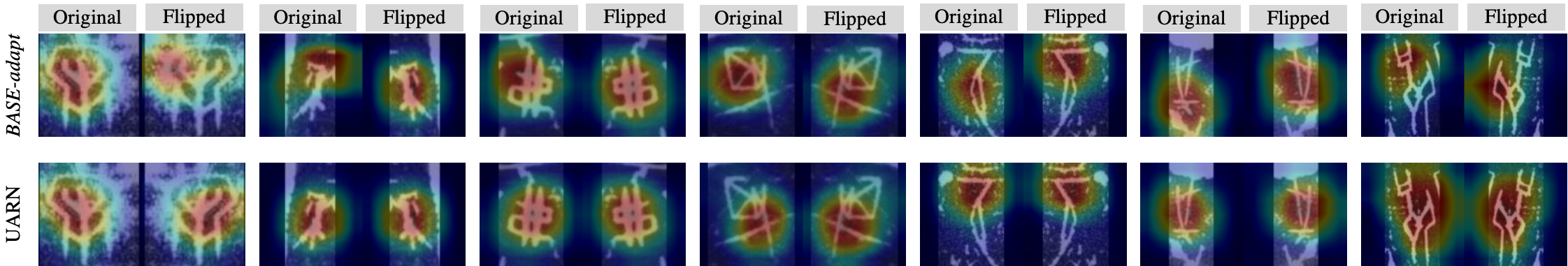}
\caption{The attention maps extracted from the original images and their flipped counterparts which are generated by \emph{BASE-adapt} and UARN.}
\label{img_visual_c} 
\end{figure*}

\begin{figure*}
\centering
\includegraphics[width=18.2cm]{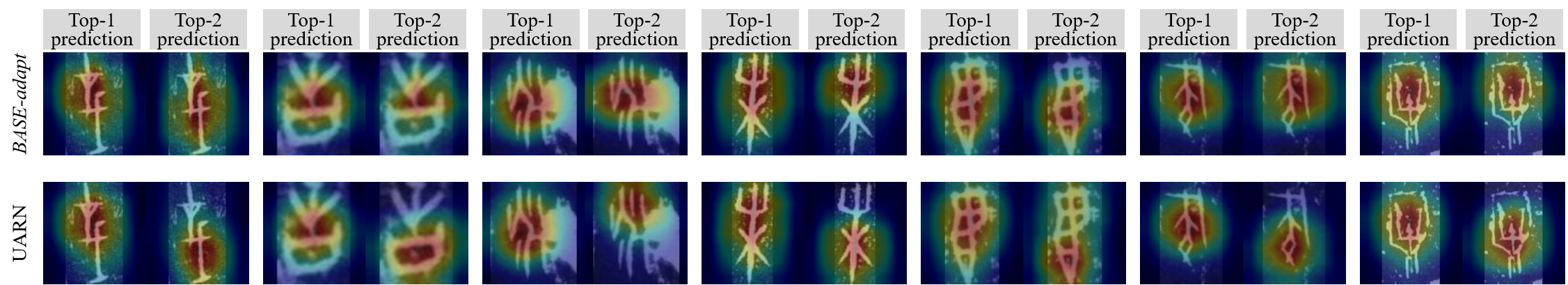}
\caption{The attention maps of the pseudo class (top-1 prediction) and the most confusing class (top-2 prediction) which are generated by \emph{BASE-adapt} and UARN.}
\label{img_visual_s} 
\end{figure*}

\begin{table*}
\renewcommand\arraystretch{1.1}
\caption{Target accuracies (mean$\pm$ std\%) on Office-31 datasets. The best accuracy is
indicated in \textbf{\textcolor{red}{bold red}} and the second best accuracy is indicated in \textcolor{blue}{\underline{underlined blue}}.}
\label{Office-31}
	\begin{center}
    \small
    \setlength{\tabcolsep}{2.5mm}{
	\begin{tabular}{l|cccccc|c}
		\toprule
         Methods &   A$\rightarrow $W  & D$\rightarrow $W & W$\rightarrow $D & A$\rightarrow $D & D$\rightarrow $A & W$\rightarrow $A & \cellcolor{gray!7} Avg \\ \hline \hline
         ResNet50 \cite{he2016deep} & 68.4$\pm$0.2 & 96.7$\pm$0.1 & 99.3$\pm$0.1 & 68.9$\pm$0.2 & 62.5$\pm$0.3 & 60.7$\pm$0.3 & \cellcolor{gray!7} 76.1 \\
         DAN \cite{long2018transferable} & 80.5$\pm$0.4 & 97.1$\pm$0.2 & 99.6$\pm$0.1 & 78.6$\pm$0.2 & 63.6$\pm$0.3 & 62.8$\pm$0.2 & \cellcolor{gray!7} 80.4  \\
         RTN \cite{Long2016Unsupervised} &  84.5$\pm$0.2 & 96.8$\pm$0.1 & 99.4$\pm$0.1 & 77.5$\pm$0.3 & 66.2$\pm$0.2 & 64.8$\pm$0.3 & \cellcolor{gray!7} 81.6 \\
         DANN \cite{Ganin2015Unsupervised} & 82.0$\pm$0.4 & 96.9$\pm$0.2 & 99.1$\pm$0.1 & 79.7$\pm$0.4 & 68.2$\pm$0.4 & 67.4$\pm$0.5 & \cellcolor{gray!7} 82.2 \\
         ADDA \cite{Tzeng2017Adversarial} &  86.2$\pm$0.5 & 96.2$\pm$0.3 & 98.4$\pm$0.3 & 77.8$\pm$0.3 & 69.5$\pm$0.4 & 68.9$\pm$0.5 & \cellcolor{gray!7} 82.9 \\
         JAN \cite{long2017deep} &  85.4$\pm$0.3 &  97.4$\pm$0.2 & \textcolor{blue}{\underline{99.8$\pm$0.2}} & 84.7$\pm$0.3 & 68.6$\pm$0.3 & \textcolor{blue}{\underline{70.0$\pm$0.4}} &  \cellcolor{gray!7} 84.3 \\
         MADA \cite{pei2018multi} &  \textcolor{blue}{\underline{90.0$\pm$0.2}} & 97.4$\pm$0.1 & 99.6$\pm$0.1 & \textcolor{blue}{\underline{87.8$\pm$0.2}} & 70.3$\pm$0.3 & 66.4$\pm$0.3 & \cellcolor{gray!7} 85.2 \\
         GTA \cite{sankaranarayanan2018generate} &  89.5$\pm$0.5 & \textcolor{blue}{\underline{97.9$\pm$0.3}} & \textcolor{blue}{\underline{99.8$\pm$0.4}} & 87.7$\pm$0.5 & \textcolor{blue}{\underline{72.8$\pm$0.3}} & \textcolor{red}{\textbf{71.4$\pm$0.4}} & \cellcolor{gray!7} \textcolor{blue}{\underline{86.5}} \\
         \hline \rowcolor{blue!5}
         \textbf{UARN (ours)} &  \textcolor{red}{\textbf{90.7$\pm$0.4}} &  \textcolor{red}{\textbf{98.6$\pm$0.2}} &  \textcolor{red}{\textbf{100.0$\pm$0.0}} & \textcolor{red}{\textbf{93.7$\pm$0.5}} & \textcolor{red}{\textbf{74.0$\pm$0.2}} & 69.2$\pm$0.6 & \cellcolor{gray!7} \textcolor{red}{\textbf{87.7}}  \\
         \bottomrule
	\end{tabular}}
    \end{center}
\end{table*}

\subsection{Visualization}

\textbf{Convergence.} To illustrate the convergence of UARN, we evaluate the source and target accuracies on Oracle-241, as shown in Fig. \ref{Convergence}. It demonstrates the efficient convergence of our UARN along the alternative iteration process. Compared with STSN \cite{9757826}, our proposed method shows a faster convergence rate and significantly lower test error on the target domain. Since STSN utilizes GAN to transform handprinted data to scanned characters, domain adaptation cannot even be achieved until the generator converges which slows down the convergence. Benefiting from attention consistency and discriminability, the optimization of our UARN is simple and stable.

\textbf{Feature visualization.} We visualize the t-SNE embeddings \cite{maaten2008visualizing} of the learned features by ResNet, DANN \cite{Ganin2015Unsupervised}, STSN \cite{9757826} and our UARN on Oracle-241 dataset. As shown in Fig. \ref{visual}, the source and target domains separate from each other for the features of ResNet. Although DANN and STSN can mix up the two domains, the features are not well-aligned. Compared with them, our UARN can achieve a better alignment. Moreover, we also visualize the target features to verify the effectiveness of UARN on improving model discriminability. We randomly select some target images belonging to 60 classes from Oracle-241, and show their t-SNE embeddings in Fig. \ref{visual_target}. ResNet and DANN both fail to classify target samples well, while STSN and our UARN learn more discriminative features on the target domain. Our UARN applies a simpler constraint, achieving comparable and even better class separation compared with STSN.

Similar observation can be observed on the U$\rightarrow $M and S$\rightarrow $M tasks of digit datasets as shown in Fig. \ref{visual_usps2mnist} and \ref{visual_svhn2mnist}. Compared with ResNet and DANN \cite{Ganin2015Unsupervised}, the source and target samples, adapted by our UARN, are well aligned. Besides, the clear separation of target samples across different categories, even in the more challenging S$\rightarrow $M task, demonstrates the robustness and discriminative ability of our learned model.

\textbf{Image visualization.} To verify that the attention maps are reﬁned by attention consistency, we show some attention maps extracted from the original and flipped images using \emph{BASE-adapt} and our UARN in Fig. \ref{img_visual_c}. We denote UARN w/o $\mathcal{L}_{ac}+\mathcal{L}_{as}$ as \emph{BASE-adapt}. It can be observed that \emph{BASE-adapt} cannot produce consistent attention maps under the flipping transformation without the constraint of $\mathcal{L}_{ac}$. Our UARN achieves better visual perceptual plausibility and shows better consistency. Moreover, we investigate the effectiveness of attention discriminability by comparing the attention maps between the pseudo class (top-1 prediction) and the most confusing class (top-2 prediction) which are generated by \emph{BASE-adapt} and our UARN. As we can see in Fig. \ref{img_visual_s}, \emph{BASE-adapt} attends to similar regions across different classes leading to visual confusion, while our UARN successfully makes the attention map separable and tells the confusing class apart.

\textbf{Generalization on object classification dataset.} To further validate the generalizability of our UARN on other tasks, we also conduct experiments on Office-31 and show the results in Table \ref{Office-31}. Office-31 is a widely adopted UDA dataset in object classification, which contains 4,652 images in 31 categories from three domains, i.e., Amazon (A), Webcam (W) and DSLR (D). We observe that our UARN achieves the best or comparable results on six transfer tasks, and obtains an average accuracy of 87.7\% on the target domains. It demonstrates UARN has a good generalization ability even if it is designed for oracle character recognition.

\section{Conclusion}

In this paper, we propose a novel unsupervised attention regularization network (UARN) for UDA of oracle character recognition. We take interpretability into consideration and encourage better visual perceptual plausibility when adapting. To be specific, we constrain attention consistency under the flipping transformation to improve the model robustness, and simultaneously enforce attention separability between the pseudo class and the most confusing class to improve the model discrimination. Comprehensive comparison experiments on Oracle-241 and MNIST-USPS-SVHN datasets strongly demonstrate the state-of-the-art performance of our UARN, when compared with other competing approaches.

However, several limitations, including future works, need to be addressed. First, we made the assumption, as is common in many existing UDA methods, that the category distribution is balanced. However, this assumption may not always hold for oracle character recognition due to the presence of rare characters. In our future work, we will delve into addressing the class imbalance issue during the adaptation process. Second, we plan to enhance our work by integrating the intrinsic properties of oracle characters, such as radicals and components, into UARN.

{
\bibliographystyle{IEEEtran}
\bibliography{egbib}
}

\end{document}